%% file: paper_template.tex
\documentclass[conference]{IEEEtran}
\usepackage{times}

\usepackage[numbers]{natbib}
\usepackage{multicol}
\usepackage[bookmarks=true]{hyperref}
\usepackage{xcolor} 
\usepackage{booktabs}
\usepackage{amsfonts}
\usepackage{amssymb}
\usepackage{amsmath}
\usepackage{graphicx}
\usepackage{pifont}
\usepackage{amsthm}

\newcommand{\cmark}{{\color{green}\ding{51}}}%
\newcommand{\xmark}{{\color{red}\ding{55}}}%

\newtheorem{theorem}{Theorem}

\begin{document}


\title{Neural Inertial Odometry from Lie Events}

\author{Royina Karegoudra Jayanth\quad Yinshuang Xu\quad Evangelos Chatzipantazis\quad Kostas Daniilidis\quad Daniel Gehrig\\Department of Computer Science, University of Pennsylvania\\
Philadelphia, PA 19104, USA \\
\texttt{\{royinakj,xuyin,vaghat,kostas,dgehrig\}@seas.upenn.edu}}



%
\newcommand{\RK}[1]{\textcolor{purple}{[Royina: #1]}}
\newcommand{\DG}[1]{\textcolor{red}{[Daniel: #1]}}
\newcommand{\YX}[1]{\textcolor{green}{[Yinshuang: #1]}}
\newcommand{\EC}[1]{\textcolor{grey}{[Evangelos: #1]}}
\newcommand{\KD}[1]{\textcolor{blue}{[Kostas: #1]}}

\maketitle

\input{0_abstract}

\IEEEpeerreviewmaketitle

\input{1_intro} 
\input{2_related}
\input{3_method}

\input{4_experiment}
\input{5_conclusion}

\bibliographystyle{plainnat}
\bibliography{references}
\newpage
\input{X_suppl}

\end{document}

%% file: 0_abstract.tex
\begin{abstract}
Neural displacement priors (NDP) can reduce the drift in inertial odometry and provide uncertainty estimates that can be readily fused with off-the-shelf filters. However, they fail to generalize to different IMU sampling rates and trajectory profiles, which limits their robustness in diverse settings. 
To address this challenge, we replace the traditional NDP inputs comprising raw IMU data with \textit{Lie events} that are robust to input rate changes and have favorable invariances when observed under different trajectory profiles. Unlike raw IMU data sampled at fixed rates, Lie events are sampled whenever the norm of the IMU pre-integration change, mapped to the Lie algebra of the $SE(3)$ group, exceeds a threshold. Inspired by event-based vision, we generalize the notion of level-crossing on 1D signals to level-crossings on the Lie algebra and generalize binary polarities to normalized \textit{Lie polarities} within this algebra. We show that training NDPs on Lie events incorporating these polarities reduces the trajectory error of off-the-shelf downstream inertial odometry methods by up to 21\% with only minimal preprocessing. We conjecture that many more sensors than IMUs or cameras can benefit from an event-based sampling paradigm and that this work makes an important first step in this direction.
\end{abstract}
\section*{Multimedia Material}
Open source code can be found here: \url{https://github.com/RoyinaJayanth/NIO_Lie_Events}.

%% file: 1_intro.tex
\section{Introduction}
\label{sec:intro}
Visual inertial odometry (VIO) has become a staple of modern localization and navigation systems powering a diverse range of applications including Augmented and Virtual Reality (AR/VR)~\cite{chen2021rnin}, autonomous driving, and robotics~\cite{Geneva20icra}.
In short, it works by integrating accelerometer and gyroscope measurements from an inertial measurement unit (IMU) and correcting the resulting drift with observations from a standard frame camera~\cite{Geneva20icra}. 
However, the usefulness of these visual observations is often limited by the quality of the captured camera frames, which degrades significantly, especially in challenging lighting conditions and high-speed motion scenarios.

\input{events_overview}

In seeking to overcome these limitations, a promising alternative has emerged, namely using \emph{neural displacement priors} (NDPs). NDPs learn to map raw IMU measurements themselves to displacement and covariance terms, and these can then be used to correct drift, while circumventing the pitfalls of classical frame-based algorithms. Surprisingly, these priors have been shown to possess similar drift-correcting capabilities to frame-based observations~\cite{liu2020tlio}, and have thus resulted in a resurgence of IMU-only, \emph{i.e.} purely \emph{Inertial} Odometry (IO)~\cite{liu2020tlio,herath2020ronin,Cao22cvpr}. 
These neural priors generate denoised displacement measurements with associated uncertainties by recognizing patterns in the IMU data and can be readily fused using off-the-shelf filters. 
However, \emph{learning} generalizable priors has proven to be a challenging endeavor. 
This is because IMU data exhibits a high degree of variability as a result of differing IMU mount orientations, motion direction, and \emph{motion patterns}. 
While orientation variability can be addressed via data augmentation~\cite{liu2020tlio}, consistency losses~\cite{Cao22cvpr} or equivariance~\cite{jayanth2024eqnio}, addressing the variability in motion patterns remains elusive.
In particular, NDPs need to learn to ignore data variability which may arise from different gaits or speeds. Yet, modeling this variation for the purposes of enforcing consistency remains an open challenge.

In this work, we take on this challenge by training NDPs with \emph{Lie Events} instead of raw IMU data, visualized in Fig.~\ref{fig:method_overview}. Lie Events are generated when the change in IMU pre-integration exceeds a pre-specified threshold. IMU pre-integrations correspond to raw integrations of debiased IMU accelerometer and gyroscope measurements. Change is measured by projecting the endpoint of the pre-integrated path onto the Lie algebra of $SE(3)$ and taking its norm. We show both theoretically and empirically that this behavior imbues the resulting events with favorable invariances with respect to different sampling rates and trajectory profiles, enhancing the robustness of downstream NDPs. 
To generate these events, we take inspiration from classical event-based vision~\cite{Lichtsteiner08ssc,Posch08iscas}, and generalize notions of level-crossing and event polarities to arbitrary Lie groups. In particular, we generalize level-crossing on 1D signals to level-crossing in the Lie algebra centered at a reference element and introduce \emph{Lie polarities} which generalize binary polarities to normalized elements in this Lie algebra. They characterize the direction of significant pre-integration change in the Lie algebra. Unlike their binary counterparts, Lie polarities are far richer since they can encode analog values. These generalizations reduce to the standard case in ~\cite{Lichtsteiner08ssc, Posch08iscas} for 1D signals. Finally, we show that event generation partially canonicalizes the IMU data with respect to motion variations within an interval. This canonicalization transforms IMU sequences to a space with lower data variability and thereby enhances the generalizability of NDPs trained in this space. Our contributions are the following:

\begin{itemize}
    \item We present a canonicalization scheme for IMU data which converts them into Lie Events. These Lie Events are directly compatible with a wide range of existing NDPs, and training on Lie Events enhances their robustness with only minimal preprocessing.
    \item We show both empirically and theoretically that these Lie Events possess favorable invariances with respect to motion variations within an interval, highlighting their benefit for NDP training. 
    \item Finally, we derive the formalism necessary to generate Lie Events, which requires the extension of traditional notions of level-crossing and event polarities to arbitrary Lie groups. These tools are applicable in a wide range of settings, beyond IMU-based odometry.
\end{itemize}

To show the generality of our method we apply it to multiple down-stream NDPs and multiple datasets, where we show consistent error reductions, with only minimal preprocessing. 

%% file: events_overview.tex
\begin{figure}[bt!]
    \centering
    \includegraphics[width=1\linewidth]{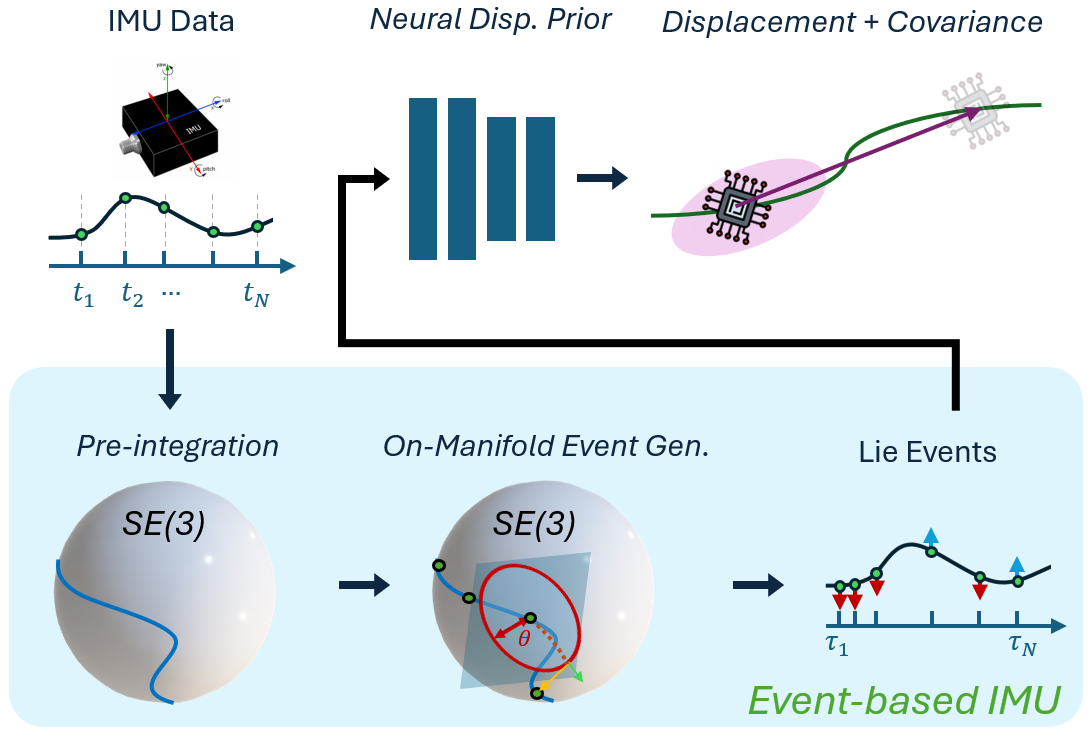}
    \caption{Neural Inerial Odometry from Lie Events. We train Neural Displacement Priors (NDPs), which enable low-drift inertial odometry, with \emph{Lie Events} derived from acceleration $\boldsymbol{a}(t_i)$ and angular rate $\boldsymbol{\omega}(t_i)$ measurements from an Inertial Measurement Unit (IMU). These events enhance the robustness of NDPs due to their favorable properties under varying sampling rates and trajectory profiles. To generate these events, we produce pre-integrations $\mathbf{x}(t)$ which reside in the special Euclidean group $SE(3)$ and then perform level-crossing on this signal which prompts the generalization of level-crossing and event polarities (red and blue arrows) to higher dimensional manifolds.}
    \label{fig:method_overview}
\end{figure}

%% file: 2_related.tex
\vspace{-3mm}
\section{Related Work}

\textbf{Neural Prior for IO:} Learning-based inertial odometry (IO) effectively reduces integration drift by leveraging neural priors. Recent approaches~\citep{Yan_2018_ECCV, pdrnet, herath2020ronin} introduce neural priors via velocity regression. Specifically, RIDI~\citep{Yan_2018_ECCV} utilizes CNNs to predict velocity for IMU data refinement, PDRNet~\citep{pdrnet} applies recurrent networks for velocity regression, and RONIN~\citep{herath2020ronin} employs TCNs to directly integrate the predicted velocity for neural-based fusion. Some works~\citep{chen2018ionet,liu2020tlio,chen2021rnin,sun2021idol} learn the displacement prior. Specifically, TLIO~\citep{liu2020tlio}  predicts both $3D$ displacements and diagonal covariances via a network, and incorporates them in an Extended Kalman Filter (EKF) via a measurement equation. RNIN-VIO~\citep{chen2021rnin} extends TLIO to continuous pedestrian motion estimation with visual inputs. In the meanwhile, there are some methods  ~\citet{9119813,deepimubias,aiimu} that leverage the power of deep learning to denoise the motion and~\citet{9811989} predicts the IMU biases. Our proposed Lie events can be directly applied to  TLIO~\citep{liu2020tlio}, and enhance its robustness to variations in motion and sampling rate.

\textbf{Canonicalization:} Standardizing inputs~\citep{lowe2004distinctive, yuceer1993rotation, jaderberg2015spatial} has long been widely practiced with both hand-crafted and learned techniques. Recently, canonicalization with equivariant models to improve generalization in off-the-shelf models has gained popularity. \citet{kaba2023equivariance} first theoretically proposed using Learned Canonicalization Functions to achieve equivariance with general off-the-shelf models. In practical applications, \citet{li2021closer,baker2024explicit} canonicalize point clouds by constructing global and local frames, \citet{deng2021vector} introduces an equivariant canonicalization method in PointNet for 3D shape analysis and reconstruction, and \citet{jayanth2024eqnio} learns an equivariant frame to canonicalize IMU data with respect to reflections and rotations. Additionally, with the rise of large pretrained models, recent work~\citep{mondal2023equivariant} applies canonicalization as an input preprocessing step to such models.

\textbf{Event-based Sensors:} Event-based sensors are most prominently used in the computer vision community, where they address the bandwidth-latency tradeoff~\cite{Forrai23icra,Gehrig24nature} of algorithms based on fixed-rate images from standard frame-based image sensors. Capturing high-speed motion requires high framerates, and, as a result bandwidth, and reducing this framerate risks missing vital scene dynamics due to the increased latency. Event cameras like the dynamic vision sensor (DVS)~\cite{Lichtsteiner08ssc}, address this by adapting their sampling rate to the scene dynamics. They do this by responding only to changes, \emph{i.e.} performing event-based sampling on the log intensity signal observed at a single pixel. Alternative cameras, like the Asynchronous Time-based Image Sensor (ATIS)~\cite{Posch08iscas}, use the same event-based sampling technique but instead measure absolute intensities. Most recently, the work in \cite{Sundar24cvpr} introduced the generalized event camera, which explored a large range of possible event cameras differing in the condition when an event is triggered, and what type of data is read out for each event. Beyond the vision domain, event-based sensing has been also generalized to asynchronous binaural spatial audition sensors~\cite{Liu14tbcas}, or tactile sensors~\cite{Funk24tro}. A proof of concept presented in ~\cite{Muglikar21threedv} shows the benefits of using event-guided depth-sensing, where an event camera is used to provide guidance to a structured light system for efficient scene reconstruction. This growing set of event-based sensors inspires the development of an event-based IMU, presented in this work. Similar to works in event-based control theory~\cite{Jin20automatica} which consider attitude consensus, our work focuses on generating events on the manifold. However, unlike~\cite{Jin20automatica}, we use these events for inertial odometry and consider the full $SE(3)$ manifold instead of only $SO(3)$. 

%% file: 3_method.tex
\section{Method}
\label{method}
The goal of this work is neural inertial odometry from a single IMU, comprised of an accelerometer providing linear acceleration measurements $\bar{\boldsymbol{a}}(t_i)\in\mathbb{R}^3$, and a gyroscope providing angular velocity measurements $\bar{\boldsymbol{\omega}}(t_i)\in\mathbb{R}^3$ at discrete times $t_i$.
These measurements are related to the true angular velocity $\boldsymbol{\omega}(t_i)$ and acceleration $\boldsymbol{a}(t_i)$ via 
{\small
\begin{align}
    \boldsymbol{\bar{\omega}}(t_i) &= \boldsymbol{\omega}(t_i) + \mathbf{b}^g(t_i) + \boldsymbol{\eta}^g(t_i)\\
   \quad \bar{\boldsymbol{a}}(t_i) &= \boldsymbol{a}(t_i) - {\mathbf{R}_{wb}}^{\intercal} (t_i)  \mathbf{g} + \mathbf{b}^a(t_i) + \boldsymbol{\eta}^a(t_i)
\end{align}
}
where $\mathbf{g}\in\mathbb{R}^3$ denotes the gravity direction pointing downward in world frame $w$, $\mathbf{R}_{wb}(t_i)$ is the transformation between body frame $b$ and $w$ at time $t_i$, and $\mathbf{b}^g,\mathbf{b}^a$ and $\boldsymbol{\eta}^g,\boldsymbol{\eta}^a_i$ are IMU biases and noises respectively.

As in previous works~\cite{liu2020tlio} we assume access to a gravity direction and bias estimate via a filter, with which IMU measurements are gravity-compensated and bias-corrected: 
\begin{align}
    \boldsymbol{\hat{\omega}}(t_i) &= \hat{\mathbf{R}}_{g}(t_i)(\bar{\boldsymbol{\omega}}(t_i) - \hat{\boldsymbol{b}}^g(t_i))\\
    \hat{\boldsymbol{a}}(t_i) &= \hat{\mathbf{R}}_g(t_i)(\bar{\boldsymbol{a}}(t_i) - \hat{\boldsymbol{b}}^a(t_i)) + \boldsymbol{g}
\end{align}
where $\hat{\mathbf{R}}_g(t_i)$ is the estimated gravity aligned frame, such that $\hat{\boldsymbol{R}}_g(t_i) \hat{\mathbf{R}}^{\intercal}_{wb}(t_i)\boldsymbol{g}=\hat{\mathbf{R}}_{\gamma}(t_i)\boldsymbol{g}=\boldsymbol{g}$ and $\hat{\mathbf{R}}_\gamma(t_i)$ is an unobservable yaw around the gravity axis.
The resulting measurements thus mimic the true body rates $\boldsymbol{\omega}(t_i),\boldsymbol{a}(t_i)$. Note that estimates of gravity and biases can be easily found by running an EKF~\cite{liu2020tlio}.

We address inertial odometry with \emph{neural displacement priors} that regress displacements and covariances from a sequence of IMU measurements regularly sampled at timestamps $t_i = i \Delta t$ within a time interval $\mathcal{I}=[0, T]$. We will denote these sampling times with $\mathcal{T}=(t_1,...,t_N)$ and $N=\lfloor\frac{T}{\Delta t}\rfloor$. 
We will use the shorthand $\hat{\boldsymbol{\omega}}(\mathcal{T})$ to denote the set $\{\hat{\boldsymbol{\omega}}(t_i)\}_{t_i\in\mathcal{T}}$. Then the displacement prior can be written as 
\begin{equation}
    \boldsymbol{d}, \boldsymbol{\Sigma} = \Phi(\hat{\boldsymbol{\omega}}(\mathcal{T}),\hat{\boldsymbol{a}}(\mathcal{T})).
\end{equation}
Here $\boldsymbol{d}\in\mathbb{R}^3$ is the displacement between times $t=0$ and $t=T$, and $\boldsymbol{\Sigma}\in\mathbb{R}^{3\times 3}$ is its associated covariance. Note that our exposition also encompasses methods like ~\cite{herath2020ronin} which predict averaged velocity $\mathbf{v}$ over a time window. Displacement can be simply recovered via $\mathbf{d}=\mathbf{v} T$, with constant $T$.

\input{time_dilation.tex}

In what follows, we will derive steps to make the above neural network robust to variations in IMU speed, thus enhancing its generalizability. In particular, we achieve this by generating canonicalized inputs
$\hat{\boldsymbol{\omega}}(\mathcal{E}),\hat{\boldsymbol{a}}(\mathcal{E})$ sampled at specifically chosen \emph{event timestamps} $\mathcal{E}=(\tau_1,...,\tau_M)$, and augmented by \emph{Lie polarities} $\mathbf{p}(\mathcal{E})$ with $\mathbf{p}(\tau_j)\in\mathbb{S}^5\subset \mathbb{R}^6$ via a suitable preprocessing step, using on-manifold event generation. 
These polarities are normalized vectors that encode the change direction of pre-integrations, and will be defined later. We term these network inputs \emph{Lie events}.  
In summary, our neural network uses inputs
\begin{equation}
\label{eq:network}
    \boldsymbol{d}, \boldsymbol{\Sigma} = \Phi(\hat{\boldsymbol{\omega}}(\mathcal{E}),\hat{\boldsymbol{a}}(\mathcal{E}), \boldsymbol{p}(\mathcal{\mathcal{E}})).
\end{equation}
In particular, we parametrize the uncertainty with a diagonal covariance $\boldsymbol{\Sigma}=\text{exp}\left(\text{diag}(2u_x,2u_y,2u_z)\right)$ with learnable $u_x,u_y,u_z$ which ensures positive definiteness of $\boldsymbol{\Sigma}$.
Next, we will study the impact of speed variations on the original neural network inputs, and then how to derive the new inputs above.

\input{events_time_reparametrization}
\subsection{Modeling Speed Variations in IMU motion}
Let the IMU trajectory\footnote{Subscripts $wb$ (body to world transformation) are omitted for brevity.} $\mathbf{T}(t)=(\mathbf{R}(t),\mathbf{t}(t))\in SE(3)$ on the interval $\mathcal{I}$ be decomposed into 
\begin{equation}
\mathbf{T}(t)=\mathbf{T}^*(\phi(t))    
\end{equation}
with \emph{path}\footnote{We distinguish between \emph{path} which is simply the sequence of points, and \emph{trajectory} which depends on time.} $\mathbf{T}^*(s)$ parametrized by arc-length $s$ and unit velocity $\Vert \mathbf{v}^*(s)\Vert = 1$ and \textit{time parametrization} $\phi: \mathcal{I}\rightarrow [0,1] $ with speed $\dot{\phi}(t)>0$ and $\phi(0)=0$ and $\phi(T)=1$.
Note that network outputs in Eq.~\eqref{eq:network} only depend on the path, since 
\begin{equation}
    \mathbf{d} = \mathbf{R}_{0}^\intercal(\mathbf{t}_T - \mathbf{t}_0)={\mathbf{R}^*}_{0}^\intercal(\mathbf{t}^*_1 - \mathbf{t}^*_0)=\mathbf{d}^*
\end{equation}
where $\mathbf{R}_0,\mathbf{t}_0$ and $\mathbf{R}_T,\mathbf{t}_T$ denote the pose at time $t=0$ and $t=T$ respectively. Moreover, for methods like ~\cite{herath2020ronin} which predict average velocity  $\mathbf{v}=\frac{\mathbf{d}}{T}=\frac{\mathbf{d}^*}{T}=\mathbf{v}^*$ for a fixed and given $T$. Yet, the network inputs are affected by $\phi(t)$ via the kinematic relations
\begin{align}
\label{eq:diff_eq}
\boldsymbol{\omega}(t)=\left(\boldsymbol{R}^{\intercal}(t) \dot{\boldsymbol{R}}(t)\right)^\vee      \quad \boldsymbol{a}(t)=\Ddot{\boldsymbol{t}}(t)
\end{align}
where $(.)^\vee$ maps a skew-symmetric matrix to a vector.
Using the chain rule we find that 
\begin{align}
{\boldsymbol{\omega}}(t)&=\dot{\phi}(t){\boldsymbol{\omega}^*}(\phi(t))\\
{\boldsymbol{a}}(t)&=\dot{\phi}(t)^2{\boldsymbol{a}^*}(\phi(t))+\Ddot{\phi}(t){\boldsymbol{v}}^*(\phi(t)).
\end{align}
where canonical angular rate and acceleration are defined as $\boldsymbol{\omega}^*(t)=\left({\boldsymbol{R}'}^{\intercal}(s) \boldsymbol{R}'(s)\right)^\vee$ and $\boldsymbol{a}'(s)=\boldsymbol{t}''(s)$.
Note the use of $(.)'$ to denote derivatives, as these are no longer temporal derivatives.
Thus the network inputs $\hat{\boldsymbol{\omega}}(\mathcal{T})$ and $\hat{\boldsymbol{a}}(\mathcal{T})$ depend on $\phi$ in the same form as above: First, $\phi$ modulates the function evaluation time, \emph{i.e.} shifts the evaluation time from $\mathcal{T}$ to $\phi(\mathcal{T})$, and second it modulates the magnitude of the measurement via its derivative. We perform partial canonicalization of the existing inputs and provide additional canonical inputs by introducing new event timestamps $\mathcal{E}=(\tau_1,...,\tau_M)$ and Lie polarities $\mathbf{p}(\mathcal{E})$. Crucially, these event timestamps depend on the trajectory $\mathbf{T}(t)$ taken by the IMU and, by construction, will depend on the event timestamps $\mathcal{E}^*=(\sigma^*_1, ...,\sigma^*_M)$ derived from the \emph{path} $\mathbf{T}^*(s)$ via $\mathcal{E}=\phi^{-1}(\mathcal{E}^*)$\footnote{Note that since $\dot{\phi}(t)>0$ the function $\phi$ is invertible.}. Likewise, we will construct polarities $\mathbf{p}(\tau_j)$ that only depend on the path, and are thus equal to the polarities $\mathbf{p}^*(\sigma^*_j)$ derived from the path.  As a result, the inputs to the network become 
\begin{align}
    {\boldsymbol{\omega}}(\mathcal{E})&=\dot{\phi}(\mathcal{E}){\boldsymbol{\omega}^*}(\mathcal{E}^*)\\
{\boldsymbol{a}}(\mathcal{E})&=\dot{\phi}(\mathcal{E})^2{\boldsymbol{a}^*}(\mathcal{E}^*)+\Ddot{\phi}(\mathcal{E}){\boldsymbol{v}}^*(\mathcal{E}^*)\\
\mathbf{p}(\mathcal{E})&=\mathbf{p}^*(\mathcal{E}^*)
\end{align}
simplifying the dependence on $\phi$. Note that $\mathbf{p}^*(\mathcal{E}^*),\boldsymbol{v}^*(\mathcal{E}^*),\boldsymbol{\omega}^*(\mathcal{E}^*),\boldsymbol{a}^*(\mathcal{E}^*)$ are \emph{canonicalized} \textit{i.e.} only depend on path $\mathbf{T}^*(s)$, not $\mathbf{T}(t)$, and that the polarity is completely independent of $\phi$. Thus networks trained on these quantities exhibit better generalization to variation in motion $\phi$. Next, we discuss how to construct $\mathcal{E}$ and $\mathbf{p}(\tau_j)$ with the properties outlined above.

\subsection{Generating IMU Events}
To construct the event timestamps $\mathcal{E}$ we take inspiration from event-based cameras~\cite{Gallego20pami}. These cameras have independent pixels that trigger an event whenever the difference of the log intensity $\text{log}(I(t))$ at a specific pixel with respect to some reference $\text{log}(I_\text{ref})$ exceeds a threshold $\theta$. We generalize this notion by identifying the pixel intensity with a new variable termed reference signal $\mathbf{x}(t)\doteq I(t)$. Thus, given that an event was triggered at the last time step $\tau_{j-1}$ and reference $\mathbf{x}_{\text{ref},j-1}$, the next event timestamp and reference can be written as
\begin{equation}
\label{eq:event_generation_1d}
    \tau_j = \min_{t>\tau_{j-1}}\{t : \theta \leq \left\Vert \text{log}(\mathbf{x}(t)) - \text{log}(\mathbf{x}_{\text{ref},j-1})\right\Vert\}, 
\end{equation}
and $\mathbf{x}_{\text{ref},j} = \mathbf{x}(\tau_j)$.
Furthermore, event cameras report a polarity $\mathbf{p}(\tau_j)$ which can be defined as 
\begin{equation}
\label{eq:polarity}
    \mathbf{p}(\tau_j) = \frac{\mathbf{x}_{\text{ref},j}-\mathbf{x}_{\text{ref},j-1}}{\Vert \mathbf{x}_{\text{ref},j}-\mathbf{x}_{\text{ref},j-1} \Vert}\in\{-1, 1\}
\end{equation}
\emph{i.e.} the sign of the change since the last reference. Using the polarities, the references can be reconstructed via
\begin{equation}
\label{eq:rec_1d}
\mathbf{x}_{\text{ref},j}=\mathbf{x}_{\text{ref},j-1}\text{exp}(\theta\mathbf{p}(\tau_j))
\end{equation}

Setting the initial event timestamp $\tau_1=0$, and reference to $\mathbf{x}_{\text{ref},1}\doteq\mathbf{x}(\mathbf{\tau_1})$ we can generate event timestamps $\tau_1,...,\tau_M$, references $\mathbf{x}_{\text{ref},1},...,\mathbf{x}_{\text{ref},M}$, as well as event polarities $\mathbf{p}(\tau_2),...,\mathbf{p}(\tau_M)$ recursively from a given signal $\mathbf{x}(t)$. 
We will show that such a sampling scheme has the desired property $\mathcal{E}=\phi^{-1}(\mathcal{E}^*)$. In particular, as the log intensity $L(t)$ at a given pixel, we can let the reference signal $\mathbf{x}(t)$ depend on a canonical reference $\mathbf{x}(t)=\mathbf{x}^*(\phi(t))$. We can inductively prove the following theorem (see appendix, Sec. I): 

\begin{theorem}\label{th:equiv}
If $\mathbf{x}(t)=\mathbf{x}^*(\phi(t))$ with $\phi'(t)>0$ and $\phi(0)=0,\phi(T)=1$, then $\tau_j = \phi^{-1}(\sigma^*_j)$ and $\mathbf{x}_{\text{ref},j}=\mathbf{x}^*_{\text{ref},j}$, where $\sigma_j^*$ and $\mathbf{x}^*_{\text{ref},j}$ are generated from $\mathbf{x}^*(s)$.
\end{theorem}
The implications of this theorem are that for all $j$ we can simply find the event timestamps $\tau_j$ by computing $\sigma^*_j$ from the path $\mathbf{x}^*(s)$ and applying $\phi^{-1}$. Note that, since $\mathbf{x}_{\text{ref},j}=\mathbf{x}^*_{\text{ref},j}$, the polarity $\mathbf{p}(\tau_j)$, defined in Eq.~\ref{eq:polarity}, only depends on events on the path $\mathbf{x}^*(s)$, \emph{i.e.} $\mathbf{p}(\tau_j)=\mathbf{p}^*(\sigma^*_j)$. 
We visualize these relations in Fig.~\ref{fig:enter-label}.
This theorem establishes the core invariance property of Lie polarities (dependent on $\mathbf{x}_{\text{ref},j}$) to variations in parametrization $\phi(t)$. Moreover, it shows that we can derive such polarities without needing to directly access unobservable $\phi(t)$ and $\mathbf{x}^*(s)$ by simply performing event-based sampling on observable $\mathbf{x}(t)$. Crucially, we never explicitly compute $\mathbf{x}^*(s)$ or $\phi(t)$.  
Having established these event properties, we turn to finding a suitable reference signal $\mathbf{x}(t)$ for IMU inertial odometry. We will show that the codomain of this reference is $SE(3)$, and thus, we extend the notion of event-based sampling to manifolds.

\subsection{On-manifold Event Generation}
Several options exist for selecting the reference signal as long as it can be written in terms of $\mathbf{x}^*(s)$ as $\mathbf{x}(t)=\mathbf{x}^*(\phi(t))$. This excludes acceleration- or velocity-like signals since these transform as $\boldsymbol{a},\boldsymbol{\omega}$ (i.e. depend on $\dot{\phi},\Ddot{\phi}$), but includes distance-like signals, \emph{e.g.} the distance over time from some selected origin. We use relative IMU pose estimates over the time interval, which can be approximated via IMU pre-integrations $\mathbf{x}(t)=(\tilde{\mathbf{R}}(t), \tilde{\mathbf{t}}(t))$. These terms are the on-manifold forward Euler integrated  solution $\tilde{\mathbf{R}}(t),\tilde{\mathbf{t}}(t)$ to \eqref{eq:diff_eq}, using raw IMU measurements $\bar{\boldsymbol{a}}(t_i),\bar{\boldsymbol{\omega}}(t_i)$. ~\citet{Forster17troOnmanifold} provides formulae for these, given initial pose $(\tilde{\mathbf{R}}(t_1),\tilde{\mathbf{t}}(t_1))=(\hat{\mathbf{R}}(t_1),\hat{\mathbf{t}}(t_1))$ and velocity $\tilde{\mathbf{v}}(t_1)=\mathbf{v}_0$, namely 
\begin{align}
    \tilde{\mathbf{R}}(t_{i+1}) &= \tilde{\mathbf{R}}(t_{i})\text{Exp}(\{\bar{\boldsymbol{\omega}}(t_i)-\hat{\mathbf{b}}^g(t_1)\}\Delta t)\\
    \tilde{\mathbf{v}}(t_{i+1}) &= \tilde{\mathbf{v}}(t_{i}) + \tilde{\mathbf{R}}(t_{i})(\bar{\boldsymbol{a}}(t_i)-\hat{\mathbf{b}}^a(t_1))\Delta t + \mathbf{g}\Delta t\\
    \tilde{\mathbf{t}}(t_{i+1}) &= \tilde{\mathbf{t}}(t_{i}) + \tilde{\mathbf{v}}(t_{i})\Delta t +  \frac{1}{2}\mathbf{g}\Delta t^2\\\nonumber&+ \frac{1}{2}\tilde{\mathbf{R}} (t_{i})(\bar{\boldsymbol{a}}(t_i)-\hat{\mathbf{b}}^a(t_1))\Delta t^2 
\end{align}
which provides finite samples $\tilde{\boldsymbol{R}}(t_i),\tilde{\boldsymbol{t}}(t_i)$ for $t_i\in\mathcal{T}$. We define the continuous signal $\mathbf{x}(t)$ by interpolating between these samples. In particular, between samples $i$ and $i+1$ we construct the geodesic path defined on $[t_i, t_{i+1}]$ as
\begin{equation}
    \mathbf{x}(t) = \mathbf{x}(t_i)\text{Exp}\left(\frac{t - t_i}{t_{i+1}-t_i} \text{Log}(\mathbf{x}^{-1}(t_{i})\mathbf{x}(t_{i+1}))\right)
\end{equation}
Here $\text{Log}: SE(3)\rightarrow \mathfrak{se}(3)$ is the log map, which sends relative poses to elements in the tangent space, and $\text{Exp}: \mathfrak{se}(3)\rightarrow SE(3)$ is the (inverse) exponential map. In Sec.~\ref{sec:rate_sensitivity}, we show that event generation is robust to IMU rate variations, and interpolation errors.

While during test time we use the EKF state to set $\mathbf{v}_0$, during training we use ground truth, and add random perturbations to $\mathbf{v}_0$ to simulate filter uncertainty. Since pre-integrations mimic the true pose of the IMU, they approximately depend on pre-integration \emph{paths} $\tilde{\boldsymbol{R}}^*(s),\tilde{\boldsymbol{t}}^*(s)$ via 
\begin{align}
\tilde{\boldsymbol{t}}(t)=\tilde{\boldsymbol{t}}^*(\phi(t))\quad\tilde{\boldsymbol{R}}(t)=\tilde{\boldsymbol{R}}^*(\phi(t)),
\end{align}
While previously our reference signal was simply a 1-D signal (log intensity), the above pre-integrations naturally reside in $SE(3)$. Thus notions such level-crossing, and polarities need to be generalized. We start off by replacing the notion of signal change between $\mathbf{x}_{\text{ref},j-1}$ and $\mathbf{x}(t)$ with the geodesic distance in $SE(3)$, and use this to replace the $L_2$-norm in Eq.~\ref{eq:event_generation_1d} 
\begin{equation}
\label{eq:ev_lie}
    \tau_j = \min_{\tau>\tau_{j-1}}\left\{\tau :  \theta \leq \left\Vert\text{Log}\left(\mathbf{x}_{\text{ref},j-1}^{-1}\mathbf{x}(t)\right)\right\Vert \right\}
\end{equation}
and $\mathbf{x}_{\text{ref},j}=\mathbf{x}(\tau_j)$. Note that this equation mimics Eq. (34) in ~\cite{Jin20automatica} applied to event-based attitude control on $SO(3)$. We extend this to $SE(3)$ and introduce the notion of event polarities in this space, which we use to enhance the robustness of our method. 
We visualize the event generation process in Fig. ~\ref{fig:event_generation} (a). We see that on the manifold, event generation amounts to finding the timestamp when the signal projected to the tangent space centered at $\mathbf{x}_{\text{ref},j-1}$ steps outside of a ball of dimension $6$. At the point where this crossing happens we record the unit vector perpendicular to this sphere and define it as being the \emph{Lie polarity} of the event at time $\tau_j$: 
\begin{equation}
\label{eq:pol_lie}
    \mathbf{p}(\tau_j) = \frac{\text{Log}(\mathbf{x}_{\text{ref},j-1}^{-1}\mathbf{x}_{\text{ref},j})}{\Vert\text{Log}(\mathbf{x}_{\text{ref},j-1}^{-1}\mathbf{x}_{\text{ref},j})\Vert}\in\mathbb{S}^5
\end{equation}
where $\mathbb{S}^5$ denotes the 6D unit sphere.
we can reconstruct the original signal with $\mathbf{p}(\tau_j)$ recursively 
\begin{equation}
\label{eq:rec_lie}
    \mathbf{x}_{\text{ref},j} = \mathbf{x}_{\text{ref},j-1}\text{Exp}(\theta \mathbf{p}(\tau_j)).
\end{equation}
These closely mimic the formulae for event generation in Eqs.~\eqref{eq:event_generation_1d} and \eqref{eq:polarity}, but are valid for arbitrary Lie groups. Moreover, when $\mathbf{x}(t)\in \mathbb{R}$, Eqs.~\eqref{eq:ev_lie},\eqref{eq:pol_lie}, and \eqref{eq:rec_lie} above reduce to Eqs. ~\eqref{eq:event_generation_1d},\eqref{eq:polarity}, and \eqref{eq:rec_1d} and polarities $\mathbf{p}(\tau_j)\in\mathbb{S}^0=\{-1,1\}$. Note also that Theorem~\ref{th:equiv} can be seamlessly transferred to this extension, as is proven in the appendix, Sec. I.

In Fig.~\ref{fig:event_generation} (b) and (c) we illustrate the invariance of polarities to time reparametrizations by considering a toy example of a dot moving in $\mathbb{R}^2$ without orientation. We visualize two reference signals $\mathbf{x}_1(t)=\mathbf{x}^*(\phi_1(t))$ and $\mathbf{x}_2(t)=\mathbf{x}^*(\phi_2(t))$, with different speed profiles. The tangent planes around reference poses in $\mathbb{R}^2$ can be visualized as planes perpendicular to time, and the $\theta$ ball can be visualized as a cylinder parallel to time. Polarities are 2D unit vectors perpendicular to the cylinder. Projecting this image onto the $xy$-plane (Fig.~\ref{fig:event_generation} (c)) reveals that the polarities only depend on the path $\mathbf{x}^*(s)$, not $\phi$.

\subsection{Network Inputs}
As our method generates a variable number of events it is not directly compatible with off-the-shelf neural displacement priors~\cite{liu2020tlio} which are designed to handle a fixed number of IMU measurements. Inspired by event-based vision, we use event stacks ~\cite{Mostafavi19cvpr}. We convert Lie events $\boldsymbol{\hat{\omega}}(\mathcal{E})$,$\boldsymbol{\hat{a}}(\mathcal{E}), \mathbf{p}(\mathcal{E})$ 
to tensors $E_{\boldsymbol{a},\boldsymbol{\omega}},E_{\mathbf{p}}\in \mathbb{R}^{B\times 6}$  with $B=200$ via: 
\begin{align}
\label{eq:event_stack}
    E_{\boldsymbol{a},\boldsymbol{\omega}}[b] &= \frac{1}{n_b}\sum_{j=1}^M [\boldsymbol{\hat{a}}(\tau_j)\Vert \boldsymbol{\hat{\omega}}(\tau_j)]\delta[b-j^*]\\
    E_{\boldsymbol{p}}[b] &= \frac{1}{l_b}\sum_{j=1}^M \boldsymbol{p}(\tau_j)\delta[b-j^*]
\end{align}
where $M$ is the number of events triggered on the interval $[0,T]$, and $\delta$ is the Kronecker delta. Note that $M$ can be much smaller than $B$, leading to sparsity in the inputs. Future work could tackle leveraging this sparsity for efficient processing. Acceleration and angular rates have been concatenated. Here $n_b$ denotes the number of samples mapped to index $b$ and $l_b$ denotes the norm of the vector at index $b$ so as to normalize the polarity. Furthermore, $j^*=\left\lfloor\frac{j-1}{M-1}(B-1)\right\rfloor$ maps the index of the event into a range $[0, B-1]$. Note that we linearly interpolate $\hat{\boldsymbol{\omega}},\hat{\boldsymbol{a}}$ to the event timestamp. We concatenate $E_{\boldsymbol{a},\boldsymbol{\omega}}$ and $E_\mathbf{p}$ along the channel dimension, resulting in an input tensor of shape $B\times 12$.

\noindent \textbf{Note on Timestamps:} We do not feed in raw time stamps into the network since they vary with the unobservable speed $\phi(t)$ from the trajectory, not the path $\mathbf{x}^*(s)$.
They would thus re-introduce data variability, which the network would need to learn to ignore. 
We believe the network can recover the displacement by leveraging event polarities. Eq ~\eqref{eq:rec_lie} shows that
the path $\mathbf{x}^*(s)$, which is sufficient to find, $\mathbf{d}$ can be recovered from polarities alone and without raw time stamps. We argue
that the network learns to correct this reconstruction with priors and additional information from $\mathbf{a}(\tau_j)$ and $\boldsymbol{\omega}(\tau_j)$.

%% file: time_dilation.tex
\begin{figure}
    \centering
    \includegraphics[width=\linewidth]{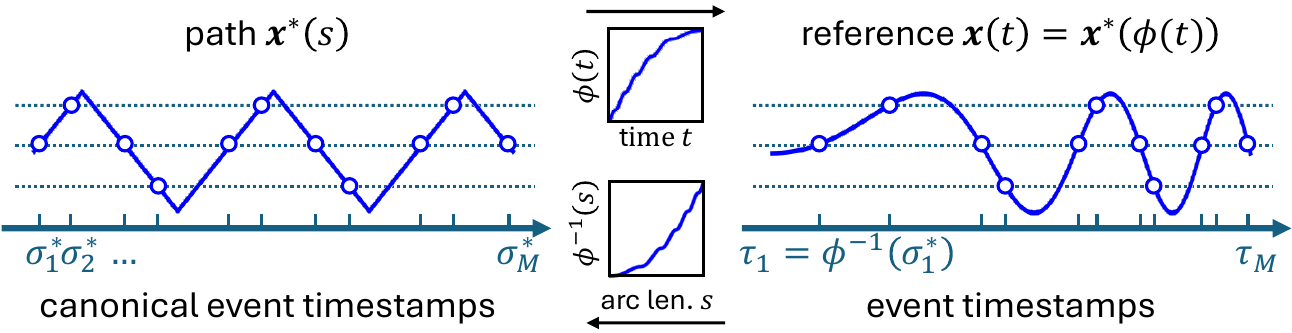}
    \caption{Event timestamps $\tau_j$ generated from reference signal $\mathbf{x}(t)=\mathbf{x}^*(\phi(t))$ can be computed from the event timestamps $\sigma^*_j$ of the path $\mathbf{x}^*(s)$, by applying the inverse mapping $\phi^{-1}$. Moreover, references $\mathbf{x}_{\text{ref},j}=\mathbf{x}(\tau_j)$ and $\mathbf{x}^*_{\text{ref},j}=\mathbf{x}^*(\sigma^*_j)$ are equal and independent of a specific $\phi$.}
    \label{fig:enter-label}
\end{figure}

%% file: events_time_reparametrization.tex
\begin{figure*}
    \centering
    \includegraphics[width=0.9\linewidth]{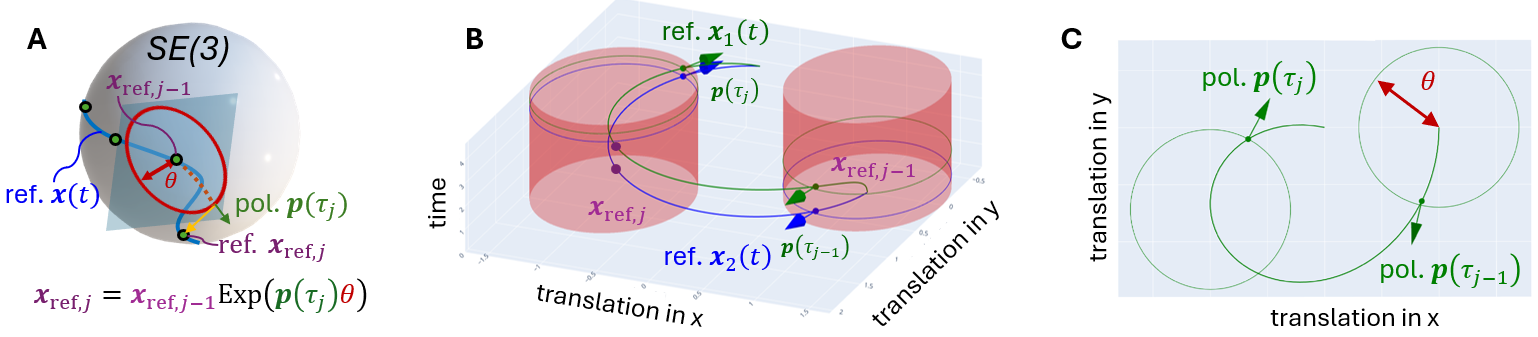}\vspace{-1ex}
    \caption{Illustration of on-manifold event generation. (A) For a reference signal $\mathbf{x}(t)$ on $SE(3)$ we find the tangent space at the last reference pose $\mathbf{x}_\text{ref,i}$. In this space, we find the moment it exits the $\theta$-ball and record the \emph{polarity} $\mathbf{p}(\tau_j)$ with unit norm perpendicular to that ball, and update the reference pose to $\mathbf{x}_{\text{ref},j}=\mathbf{x}_{\text{ref},j-1}\text{Exp}(\mathbf{p}(\tau_j)\theta)$. (B) Two trajectories $\mathbf{x}_1(t)=\mathbf{x}^*(\phi_1(t))$ and $\mathbf{x}_2(t)=\mathbf{x}^*(\phi_2(t))$ on the simplified manifold $\mathbb{R}^2$ with time parametrizations $\phi_1,\phi_2$. (C) After projecting both trajectories onto the $xy$-plane, the reference signals $\mathbf{x}_{\text{ref},j},\mathbf{x}_{\text{ref},j-1}$ and polarities $\mathbf{p}(\tau_j),\mathbf{p}(\tau_{j-1})$ are equal. }\vspace{-2ex}
    \label{fig:event_generation}
\end{figure*}

%% file: 4_experiment.tex
\section{Experiments}
First, we present a toy experiment to validate the result of Theorem~\ref{th:equiv} in Sec.~\ref{sec:toy_example}, before demonstrating the generality of our approach by applying it to the neural inertial odometry frameworks TLIO~\cite{liu2020tlio} (Sec.~\ref{sec:tlio}) and RoNIN~\cite{herath2020ronin} (Sec.~\ref{sec:ronin}). 
Furthermore, we will present a sensitivity study that highlights the robustness of our approach with respect to IMU rate variations in Sec.~\ref{sec:rate_sensitivity}, and round off with ablation and hyperparameter sensitivity studies in Sec.~\ref{sec:ablation}.

\input{toy_example.tex}
\subsection{Toy Example}
\label{sec:toy_example}
Here, we study the event timestamps $\mathcal{E}$ generated from reference signals based on the pre-integration of IMU accelerations and angular rates, and show their relation to event timestamps $\mathcal{E}^*$ generated from canonical pre-integrations. 
We use IMU data from the TLIO test dataset~\cite {liu2020tlio}, which comprises 60 hours of 1 kHz IMU data, with 200 Hz ground truth trajectory data from MSCKF~\cite{Mourikis07icra}. It was gathered from five individuals performing a broad range of activities, for example, walking, stair traversal, and kitchen organization. More dataset details are in the appendix, Sec. V.

We denote non-overlapping 1-second ground truth trajectories as paths $\mathbf{T}^*(s)$ and generate synthetic new trajectories $\mathbf{T}_\alpha(t)=\mathbf{T}^*(\phi_\alpha(t))$, where $\phi_\alpha(t)= (\frac{t}{\Delta t})^\alpha\in[0,1]$ with $\Delta t=1$s and $\alpha\in\{0.5, 1.0, 2.0\}$. We then generate artificial IMU measurements by fitting a spline through the poses and finding derivatives. We apply bias, noise, and bias drift consistent with the noise magnitudes found in ~\cite{liu2020tlio}. We then perform pre-integration on IMU data from the path $\mathbf{\hat{R}}^*(s),\mathbf{\hat{t}}^*(s)$ and trajectories $\mathbf{\hat{R}}_\alpha(t),\mathbf{\hat{t}}_\alpha(t)$. Finally, we generate events using different thresholds $\theta\in\{0.005, 0.01, 0.02\}$ producing canonical event timestamps $\mathcal{E}^*$ and non-canonical event timestamps $\mathcal{E}_\alpha$. Finally, we map $\mathcal{E}_\alpha$ to the canonical setting, i.e. compute $\phi_\alpha(\mathcal{E}_\alpha)$ and compare them with $\mathcal{E}^*$ via the chamfer distance. For comparison, we also report the chamfer distance without remapping. We compare with noiseless ground truth poses as the reference signal to show the lower error bound.  

Tab.~\ref{tab:toy_example} shows that event generation is capable of sampling the correct measurement even after time dilation. The smallest threshold $0.005$ achieves an error of 1.15 $\%$ for a remapping of $\phi(t)=t^2$. We also see that using ground truth significantly reduces this error to about $0.1\%$ in the same setting validating the theory for ideal references. Future research can benefit from more accurate pre-integration schemes. 
\input{main_table.tex}
\subsection{Application to TLIO}
\label{sec:tlio}
TLIO is a learning-based inertial odometry algorithm that estimates the 6-DoF IMU pose with an extended Kalman filter (EKF). Propagation is performed by integrating raw accelerometer and gyroscope measurements, while measurement updates are performed with neural network-based displacement and covariance predictions $\mathbf{d}\in\mathbb{R}^3$ and $\boldsymbol{\Sigma}\in\mathbb{R}^{3\times 3}$. The network takes 1-second windows of 200 IMU measurements as input with shape $200\times6$. IMU data is bias-corrected using factory calibration bias values, and gravity aligned using the EKF orientation state. EKF details and propagation and measurement equations are provided in Sec. II, and network details in Sec. IV of the appendix. 

\textbf{Implementation Details}
We pre-integrate accelerometer and gyroscope measurement sequences of length $200$ and obtain the orientation and position estimates in a gravity-aligned world frame. We use the EKF estimates of orientation, position, and initial velocity at the beginning of the time window, termed clone states, to initialize the pre-integration during testing. We then generate event timestamps and polarities using this reconstructed trajectory and a threshold $\theta=0.01$. Sec.~\ref{sec:ablation} shows a sensitivity analysis motivating the choice of $\theta$. For consistency, we map the polarity into the gravity-aligned frame after generation.

\textbf{Datasets:}
We train our method on the TLIO Dataset~\citep{liu2020tlio}, described in the appendix, and evaluate it on the TLIO Test Set and the Aria Everyday Activities (Aria) Datasets~\citep{lv2024aria}. 
Aria is an egocentric dataset collected using the Project Aria glasses~\citep{engel2023project} and comprises a left (800 Hz) and right (1 kHz) IMU with ground truth position and orientation. It contains 7.3 hours of data and includes a wide range of wearers engaged in everyday activities like reading, exercising, and relaxing. More dataset details are in the appendix, Sec. V.

\textbf{Training Details:}
We train the first 10 epochs we train with a component-wise MSE loss~\cite{liu2020tlio}, so that the displacement prediction converges, and we then, for 40 epochs, switch to a Maximum Likelihood Error (MLE) loss~\cite{liu2020tlio} which incorporates both the predicted displacement and diagonal covariance. 
We use a learning rate of $10^{-4}$, the Adam optimizer~\cite{Kingma15iclr}, and a batch size of 1024. We train all our models on NVIDIA a40 GPUs. 
During training, we use ground truth initial velocities $\mathbf{v}_0$ for bootstrapping pre-integration, and perturb it with uniform noise $\mathbf{n}_{\mathbf{v}_0}\sim\mathcal{U}[-0.5,0.5]$.
To simulate noise in the gravity-aligned frame, we perturb the gravity direction by $5^\circ$ uniformly at random and apply a random yaw. 
Furthermore, we add uniform noise  $\mathbf{n}_{\boldsymbol{\omega}}\sim\mathcal{U}[-0.05,0.05]$ and $\mathbf{n}_{\boldsymbol{a}}\sim\mathcal{U}[-0.2,0.2]$ to angular rates, and accelerations simulating uncertainties in bias estimates. 
We also add noise $\mathbf{n}_{\mathbf{p}}\sim\mathcal{U}[-0.5,0.5]$ to polarities before renormalizing.

\textbf{Baselines:} For a fair comparison, we implement three variants of TLIO~\citep{liu2020tlio}, which, similar to our method, target time scale robustness. During training, we randomly subsample the 200 Hz IMU data to rates $r\in\{20, 40, 100, 200\}$. Since natively TLIO requires an input of 200 samples, we convert the subsampled data back to 200 by interpolation (denoted \emph{TLIO + interp.}) or by using the event stack in Eq.~\eqref{eq:event_stack} (denoted \emph{TLIO + splat.}). Note that our method (denoted \emph{TLIO + events}) is not trained with rate augmentation.

\textbf{Metrics:}
We report the following metrics used in~\citep{liu2020tlio,Cao22cvpr,herath2020ronin}, and written out in detail in the appendix, Sec. VI: 
\begin{itemize}
    \item \textit{Mean Squared Error (MSE)} between predicted ($\hat{\mathbf{d}}(t_i)$) and ground truth displacements ($\mathbf{d}(t_i)$) averaged over the trajectory.
    \item Absolute Translation Error (ATE), \textit{i.e.} the RMSE between estimated ($\bar{\mathbf{t}}(t_i)$) and ground truth positions ($\mathbf{t}(t_i)$).
    \item Relative Translation Error (RTE), \textit{i.e.} local differences between $\mathbf{t}(t_i)$ and $\bar{\mathbf{t}}(t_i)$ over a 1-second or (1 minute for RoNIN) window $\delta t$.
    \item Absolute Yaw Error (AYE), \textit{i.e.} the yaw RMSE.
    \item Translational drift over the total distance traveled.
\end{itemize}
To limit the impact of outliers we report the median error over the different trajectories of the dataset. Furthermore, to evaluate the impact of the neural network and EKF separately, we evaluate two types of trajectories produced by our method: The first uses simple integration of the network outputs, and the corresponding metrics are denoted with a $*$, and the second uses the EKF, and is denoted without a $*$. \\
\textbf{Results:} We report the results in Tab.~\ref{tab:nn_main_table}, and trajectory plots in Sec. VII of the appendix. First, we see that, without EKF, events improve ATE$^*$ on all datasets. They reduce the ATE$^*$ on the TLIO Dataset by 13\%, on Aria Right by 11\%, and on Aria Left by 19\% compared to base TLIO. Moreover, it reduces MSE$^*$ on most datasets, reducing by 17\% on Aria Right, by 21\% on Aria Left, but having a  15\% higher MSE$^*$ than base TLIO on the TLIO Dataset. As in ~\cite{liu2020tlio} we argue that the correlation between MSE and ATE is not exact, since other methods may produce displacement estimates with lower variance, but higher bias, introducing larger overall drift.

Our method's EKF accuracy (ATE), is 10\% higher on the TLIO Dataset, 3\% on the Aria Right Dataset, and 12\% on the Aria Left Dataset compared to base TLIO. We also see that our method outperforms base TLIO in terms of drift but is outcompeted by methods that use data augmentation.
\subsection{Application to RoNIN}
\label{sec:ronin}
RoNIN~\cite{Herath20icra} is a 2D inertial navigation method designed for IMU-based pedestrian tracking. It works by regressing velocity predictions,$\hat{\mathbf{v}}(t)\in\mathbb{R}^2$ expressed in the local gravity aligned frame, from 1-second sequences of IMU measurements using a ResNet-18~\cite{He16cvpr}. These velocities are regressed with a 25$ms$ stride, transformed into the global frame and then integrated using known orientations from the IMU.   \\
\textbf{Implementation Details:} 
As in ~\cite{herath2020ronin} we use the RoNIN dataset, which features 200 Hz IMU data. We convert sets of 200 measurements into events, using the same parameters as for TLIO but a higher contrast threshold $\theta=0.1$ due to the higher average accelerations. We also use the last velocity prediction as initialization for pre-integration.\\
\textbf{Datasets:} We train our model on the public portion of the RoNIN dataset~\cite{herath2020ronin} which comprises 200 Hz IMU and ground truth trajectory data collected from pedestrians with varied placements and devices. We report results on the RoNIN test set, and also on the RIDI~\cite{Yan_2018_ECCV} and Oxford Inertial Odometry Dataset (OxIOD)~\cite{chen2018arxiv}. RIDI comprises pedestrian data at 200 Hz, with four different sensor placements, and includes forward, backward, and sideways walking motion as well as accelerating and decelerating motions. This dataset uses Umeyama alignment~\cite{Umeyama91pami} before calculating metrics. The OxIOD dataset also comprises smartphone-based pedestrian data with various motion modes, devices, and device placements. More dataset details are in the appendix, Sec. V.\\
\textbf{Training Details:} Training mirrors~\cite{herath2020ronin} with 50\% of the training data that is public, minimizing the robust velocity loss proposed in ~\cite{herath2020ronin}. We use ADAM~\cite{Kingma15iclr} with a batch size of 128 and an initial learning rate of $10^{-4}$ and a decay by a factor of 0.1 if the validation loss does not decrease in 10 epochs. The maximum number of epochs is 120. Furthermore, the linear layers use dropout with $p=0.5$.
\input{ronin_table}
\input{rate_exp_figure}
\input{hparam_sensitivity}
As in ~\cite{herath2020ronin} we omit gravity perturbation, and bias noise, and leave other noise sources and magnitudes the same.\\
\textbf{Baselines:} We compare our method with RoNIN operating on raw IMU data with three different temporal processing modules, namely ResNet-18~\cite{He16cvpr}, LSTM~\cite{Hochreiter97nc} and TCN~\cite{Lea16eccvw}, and also compare against RIO~\cite{Cao22cvpr}. RIO adds two additional strategies to RoNIN which enforce approximate yaw equivariance. The first one, termed joint training (denoted with \textit{+J}), optimizes an auxiliary equivariant consistency loss between predictions observed under different yaws. The other strategy, adopted during testing and termed adaptive test-time-training (denoted with \textit{+TTT}) performs gradient descent steps on the current model at 20 Hz, based on the equivariant consistency loss derived from a buffer of 128 past IMU measurements. Backpropagation is only performed on losses between samples that have a significant predicted velocity mismatch. Furthermore, test-time training and model resetting are controlled via a separate ensemble of models that estimate prediction uncertainty online. Finally, we report the results of a naive double integration baseline (denoted with \textit{Integration}).\\
\textbf{Results:} We report results in Tab.~\ref{tab:ronin}, and trajectory plots in Sec. VII for the appendix. Note that the first three rows are results reported in ~\cite{Herath20icra} which use 100\% of the training data. Other methods are retrained with the public training data, which is only 50\%. We observe that, despite this mismatch, in terms of ATE$^*$, our method outperforms RONIN on 50\% data on the RONIN dataset, on RIDI, and on OxIOD. Test-time training and yaw consistency methods can be used complementarily to our method to further improve results.\\
\textbf{Discussion: } Similar to the results when applied to TLIO, the above results highlight the strength of using events instead of regular IMU samples. Moreover, we see that our method outperforms (i) methods that were trained with more data on RIDI-T and OxIOD (see row 1-3 in Tab.~\ref{tab:ronin}) and (ii) methods that incorporate yaw consistency strategies on OxIOD. In particular, using yaw consistency strategies is an orthogonal approach to improving inertial odometry, which would additionally strengthen our result. Finally, we note that our method may benefit from a weak autoregression form by using the predicted velocity from the previous timestamp for pre-integration. However, compared to models that also perform autoregression, \textit{i.e.} RoNIN+TCN or RoNIN+LSTM, our method achieves a lower ATE$^*$.  

\subsection{Rate Sensitivity Study}
\label{sec:rate_sensitivity}
In this section, we study the robustness of our method to varying IMU measurement rates which simulate differing time scales. We generate lower-rate IMU data by subsampling the TLIO data (200Hz) to $r=\{20,40,100,200\}$, the Aria Left data (800Hz) to $r=\{20,40, 100, 200, 400, 800\}$ and the Aria Right data (1000Hz) to $r=\{20,40, 100, 200, 500, 1000\}$. This subsampling strategy tests the robustness of our event generation method to sampling errors. For the TLIO baselines (base, +interp. and +splat.) we convert the IMU data to an equivalent of 200 inputs per 1 second of data with interpolation or event stack generation. We report the ATE$^*$ in Fig.~\ref{fig:rate_exp}, and a table with the full results in the appendix, Sec. VIII. Note that despite not being trained with data augmentation our method remains stable over differing rates, outperforming the baselines and TLIO. Most methods degrade heavily at low rates, while our method degrades gracefully.

\input{ablation_table_nn}

\subsection{Ablation Study and Hyperparameter Sensitivity Analysis}
\label{sec:ablation}
In this section, we ablate the design choices in the methodology, and show sensitivity analyses of the specific chosen hyperparameters. 
First, we study the impact of the reference signal $\mathbf{x}(t)$ and its associated manifold by comparing the ATE and ATE$^*$ on the TLIO test set. We select reference signals that monitor only the translation component $\mathbf{\tilde{t}}(t)$ of the pre-integration (labeled with ``position") and those that monitor the full pose $\mathbf{\tilde{R}}(t),\mathbf{\tilde{t}}(t)$ (labeled ``pose"). While ``position" admits only the $\mathbb{R}^3$ manifold, ``pose" admits both the direct product manifold $SO(3)\times \mathbb{R}^3$ and the semidirect product manifold $SE(3)=SO(3)\ltimes \mathbb{R}^3$. These differ in the definition of the Log and Exp map used previously. We also study the impact of \emph{Lie polarities} and the necessity of adding polarity noise. We report results in Tab.~\ref{tab:ablation}. We see that of the different considered manifolds, $SE(3)$ yields the best ATE$^*$ and ATE. This points to the fact that using the full pose and associated manifold yields the most informative events. Using polarities without noise yields very low ATE$^*$ but high ATE when deployed with the EKF. We argue that the neural network learns to overfit to specific training trajectories, and thus becomes sensitive to noise in its predictions. Adding polarity noise during training significantly reduces the ATE when deployed with the EKF.
\begin{figure}[t!]
    \centering
    \includegraphics[width=\linewidth]{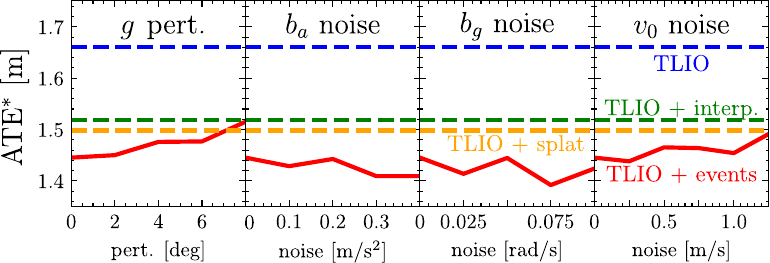}\vspace{-1.5ex}
    \caption{Sensitivity to pre-integration errors. Baseline ATE$^*$ in dashed lines.}
    \label{fig:preint}\vspace{-1.5ex}
\end{figure}
Next, we study the sensitivity of our method to hyperparameters. Specifically, we vary the contrast threshold $\theta\in\{0.005,0.1,0.2\}$, polarity noise range $r_\mathbf{p}\in\{0.1,0.25,0.5,0.75,1.0\}$ and initial velocity noise range $r_{\mathbf{v}_0}\in\{0.25,0.5,1.0,1.5\}$ applied during training and report results in Fig.~\ref{fig:hparam_sensitivity}, with tables with the full results in the appendix, Sec. VIII.
First, we find an optimal ATE at $\theta=0.01$. While too large thresholds may reduce the sensitivity of the method, too low thresholds may be more susceptible to noise and increase computational complexity. ATE reaches an optimum for initial velocity noise range $r_{\mathbf{v}_0}=0.5$. As previously discussed, using too low noise makes the network overfit to the specific trajectories and sensitive to noise, leading to excessively low ATE$^*$ but high ATE. Finally, the optimal polarity noise range is around $r_\mathbf{p}=0.75$. We found that this value yielded marginally worse results on other datasets, so we opted for $r_\mathbf{p}=0.5$.

\subsection{Runtime, Robustness and Generalization}
\noindent\textbf{Runtime:} Running TLIO for 1 second of data takes 1.14 ms on an NVIDIA GeForce RTX 4090 Laptop GPU, with an Intel i9 processor, enabling real-time measurement updates at the 20 Hz used in the experiments. The network (baseline TLIO, i.e. 1D ResNet-18, PyTorch) takes 1.06 ms. The C++-implemented event generation (including
pre-integration) takes 76 $\mu$s (4.22 $\mu$s/event), and stacking takes 7.62 $\mu$s ( 0.331 $\mu$s/event). With an average event rate of 74
Hz (vs. 200 Hz for TLIO), our method reduces redundancy.
\input{data_efficiency}

\noindent\textbf{Sensitivity to IMU Pre-integration Accuracy:} Inaccurate bias, gravity, and initial velocity estimation degrade pre-integration accuracy. 
To minimize this we train with initial velocity, gravity, and bias noise. Fig.~\ref{fig:preint} shows the sensitivity of ATE$^*$ (lower is better) with respect to gravity, bias, and initial velocity noise perturbation for pre-integration, on the TLIO test set, following the scheme in Sec.~\ref{sec:tlio} ``training details”.
These introduce drift, $SO(3)$ perturbations, and elucidate the role of gravity. Gravity and $v_0$ perturbations lead to graceful degradation, still outperforming other baselines, and bias noises show little to no degradation. Slight improvements may stem from the fact that we train with
noise. Overall this shows low impact of bias-induced drift, and
higher sensitivity to gravity and velocity estimation errors.

\begin{figure}[t!]
    \centering
    \includegraphics[width=\linewidth]{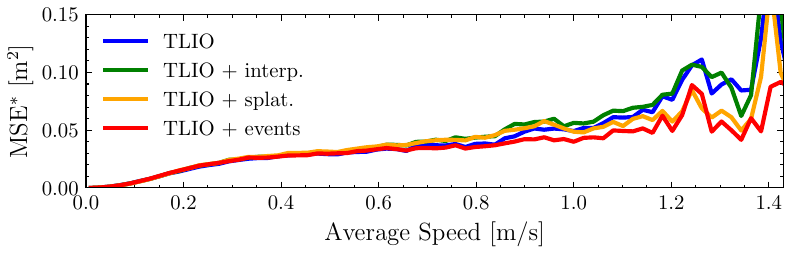}\vspace{-2ex}
    \caption{Aria dataset MSE vs. speed for models trained on the TLIO dataset.}\vspace{-3ex}
    \label{fig:aria}
\end{figure}

\noindent\textbf{Generalization:} Tab. \ref{tab:data_eff} reports our method trained on smaller subsets of the RoNIN training set. Our method's performance stays relatively stable on both in- (RoNIN-U), and out-of-distribution data (RIDI-T, OxIOD). Notably, performance on RIDI-T degrades gracefully, while fluctuating on the other datasets. Even with 10\% our method can learn reasonably well, underlining generalization, which we ascribe to the canonicalization capabilities of events.

\noindent\textbf{Out-of-distribution data: }
We report the distribution of MSE versus average speed, for models trained on the TLIO dataset and tested on the out-of-distribution Aria datasets (Fig. 2). Our method outcompetes other methods with lower MSE at higher speeds, supporting the claim that events confer canonicalization with respect to speed variations. This targets a key limitation of existing methods, and shows improvements in edge cases with large accelerations .

%% file: toy_example.tex
\begin{table}[t!]
\resizebox{\linewidth}{!}{%
\begin{tabular}{llc|ccc}
\hline
\textbf{Ref. $\mathbf{x}(t)$} & $\phi_\alpha(t)$ & \textbf{Correction}& \multicolumn{3}{c}{\textbf{Contrast Threshold}} \\ \hline
                          &                      &      & $\theta=0.005$           & $\theta=0.01$          & $\theta=0.02$          \\ \hline
pre-integration & $t^2$     & no &2.18& 3.62& 5.52 \\
               & $t^{0.5}$ & no &2.48& 3.89& 5.61 \\
               & $t^2$     & yes& 1.15&1.73& 2.45 \\
               & $t^{0.5}$ & yes& 0.95&1.11& 1.12 \\ \hline
ground truth   & $t^2$     & no &5.2& 10.4& 2.39 \\
               & $t^{0.5}$ & no &0.73& 1.37& 2.79 \\ 
               & $t^2$     & yes&0.13& 0.16& 0.21 \\
               & $t^{0.5}$ & yes&0.02& 0.03& 0.04 \\ \hline

\end{tabular}}
\caption{Chamfer distance in \% between event timestamps $\phi_\alpha(\mathcal{E})$ and canonical event timestamps $\mathcal{E}^*$. No correction denotes that $\mathcal{E}$ was not remapped using $\phi_\alpha$. }\vspace{-5ex}\label{tab:toy_example}
\end{table}

%% file: main_table.tex
\begin{table*}[ht]
\newcommand{\first}{\cellcolor{red!40}}
\newcommand{\second}{\cellcolor{orange!40}}
\newcommand{\third}{\cellcolor{yellow!40}}
\centering
\begin{small}
\resizebox{\linewidth}{!}{
\begin{tabular}{l|c|cccccc|cccccccccccc}
\toprule
&&\multicolumn{6}{c|}{\textbf{NN evaluation}}&\multicolumn{12}{c}{\textbf{NN+EKF evaluation}}\\\midrule
&& \multicolumn{2}{c}{TLIO Dataset} & \multicolumn{2}{c}{Aria Right} & \multicolumn{2}{c|}{Aria Left}&\multicolumn{4}{c}{TLIO Dataset} & \multicolumn{4}{c}{Aria Right} & \multicolumn{4}{c}{Aria Left}\\
\cmidrule(lr){3-4} \cmidrule(lr){5-6}\cmidrule(lr){7-8}\cmidrule(lr){9-12} \cmidrule(lr){13-16}\cmidrule(lr){17-20} 
Model & rate& MSE*
& ATE*& MSE* & ATE*&MSE* & ATE* & ATE & RTE & Drift & AYE & ATE & RTE & Drift & AYE & ATE & RTE & Drift & AYE\\
&aug.&($m^2$)&($m$)&($m^2$)&($m$)&($m^2$)&($m$)&($m$)&($m$)&($\%$)&($\circ$)&($m$)&($m$)&($\%$)&($deg$)&($m$)&($m$)&($\%$)&($\circ$)\\
\midrule
TLIO
&\xmark&\textbf{0.013}   &1.660&\underline{0.018}&1.240&\underline{0.019}&1.314 & \underline{1.410}&\textbf{0.102}&1.486&\underline{1.832}&0.865&0.119&2.629&1.481&1.069&\textbf{0.160}&2.749&\underline{5.203}\\
+ interp.
&\cmark&\textbf{0.013}   &1.519&0.019&\underline{1.195}&0.020&1.326 &\underline{1.410}&0.109&1.302&1.836&\underline{0.837}&0.117&\textbf{2.456}&\textbf{1.441}&\underline{0.944}&0.163&\textbf{2.335}&5.214\\
+ splat. 
&\cmark&\underline{0.014}&\underline{1.498}&0.019&1.216&0.020&\underline{1.292}&1.440&0.107&\underline{1.242}&\textbf{1.788}&0.876&\textbf{0.114}&2.479&1.462&0.965&\underline{0.161}&\underline{2.425}&\textbf{5.142}\\
\textbf{+ events (ours)}
&\xmark&0.015            &\textbf{1.445}&\textbf{0.015}&\textbf{1.099}&\textbf{0.015}&\textbf{1.064}&\textbf{1.282}&\underline{0.106}&\textbf{0.953}&1.863&\textbf{0.836}&\underline{0.115}&\underline{2.459}&\underline{1.456}&\textbf{0.942}&0.165&\underline{2.425}&5.215 \\
\bottomrule

\end{tabular}}
\caption{Comparison of our method against baselines TLIO with and without data augmentation on the TLIO Dataset~\cite{liu2020tlio}, and Aria Everyday Datasets~\cite{lv2024aria}. Note here Left and Right denote the left and right IMUs in the Aria Datasets.}
\vspace{-5ex}
\label{tab:nn_main_table}
\end{small}
\end{table*}

%% file: ronin_table.tex
\begin{table}[t!]
\newcommand{\first}{\cellcolor{red!40}}
\newcommand{\second}{\cellcolor{orange!40}}
\newcommand{\third}{\cellcolor{yellow!40}}
\centering
\begin{small}
\resizebox{\linewidth}{!}{
\begin{tabular}{lccccccc}
\toprule
&& \multicolumn{2}{c}{RONIN-U} & \multicolumn{2}{c}{RIDI-T}& \multicolumn{2}{c}{OxIOD} \\
\cmidrule(lr){3-4} \cmidrule(lr){5-6}\cmidrule(lr){7-8}
Model & Training& ATE* & RTE*& ATE* & RTE* &ATE* & RTE* \\
 &Data& (m)&(m)& (m)&(m)& (m)&(m)\\
\midrule
RoNIN + ResNet         &100\%& 5.14 & 4.37             & 1.63             & 1.91  & 3.46 & 4.39 \\
RoNIN + LSTM           &100\%& 5.46 & \underline{3.95} & 2.00             & 2.64  & 3.02 & 4.18 \\
RoNIN + TCN            &100\%& 5.87 & 4.39             & 1.66             & 1.95  & 3.56 & 4.98 \\\hline
RoNIN + ResNet         &50\% & 5.57 & 4.38             & 1.19             & 1.75 &  3.52 & 4.42 \\
+ J                    &50\% & 5.02 & 4.23             & 1.13             & 1.65 &  3.59 & 4.43\\
+ TTT                  &50\% & 5.05 & 4.14             & 1.04             & 1.53 &  2.92& 3.67\\
+ J + TTT              &50\% & 5.07 & 4.17 & 1.03 & 1.51 &2.96&3.74\\
+ \textbf{events (ours)}     &50\%&5.35&4.63&1.03&1.97&1.52&1.91\\\midrule
Integration &-&458.06&117.06&31.06&37.53&1941.41&848.55\\
\bottomrule
\end{tabular}}
\vspace{-2mm}
\caption{Application of our method to the RoNIN architecture, and comparison on the RoNIN, RIDI, and the Oxford Inertial Odometry Datasets (OxIOD). Note, * indicates trajectory reconstruction without an EKF, as reported in ~\cite{herath2020ronin,Cao22cvpr}.}\vspace{-3ex}
\label{tab:ronin}
\end{small}
\vspace{-2mm}
\end{table}

%% file: rate_exp_figure.tex
\begin{figure*}
\begin{tabular}{c}
     \includegraphics[width=\linewidth]{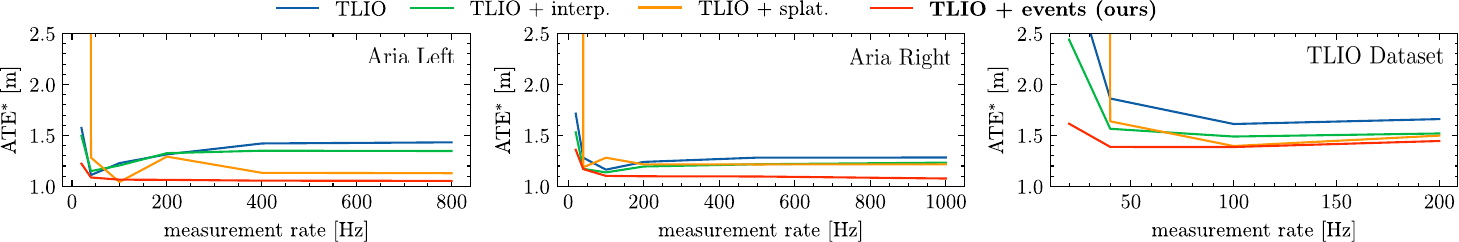}
\end{tabular}\vspace{-1ex}
    \caption{IMU rate sensitivity analysis. Each method is trained on the TLIO training set. Methods + interp. and + splat. were trained with IMU rate augmentation and TLIO and TLIO + events were trained without data rate augmentation. }\vspace{-1ex}
    \label{fig:rate_exp}
    \vspace{-2mm}
\end{figure*}

%% file: hparam_sensitivity.tex
\begin{figure*}
    \centering
     \setlength{\tabcolsep}{3pt}
    \begin{tabular}{ccc}
    \includegraphics[width=0.32\linewidth]{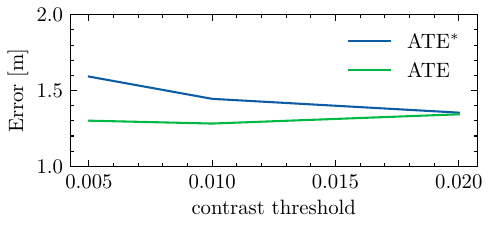}&
    \includegraphics[width=0.32\linewidth]{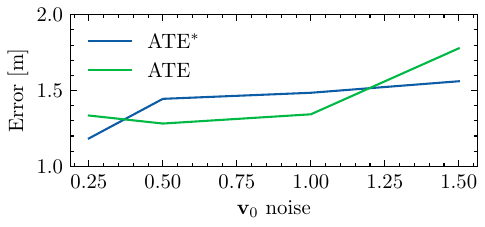}&
    \includegraphics[width=0.32\linewidth]{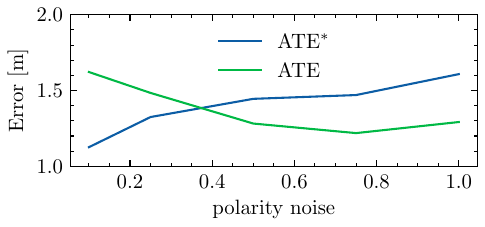}
    \end{tabular}\vspace{-1ex}
    \caption{Hyperparameter Sensitivity: (left) the contrast threshold determines the distance to the reference when events are fired, (middle) initial velocity $\mathbf{v}_0$ and (right) polarity 
 $\mathbf{p}(\tau_j)$ noise determine the range of uniform noise perturbing these quantities during training. In particular, $\mathbf{v}_0$ noise affects pre-integration.}\vspace{-2ex}
    \label{fig:hparam_sensitivity}
\end{figure*}

%% file: ablation_table_nn.tex
\begin{table}
\newcommand{\first}{\cellcolor{red!40}}
\newcommand{\second}{\cellcolor{orange!40}}
\newcommand{\third}{\cellcolor{yellow!40}}
\centering
\begin{small}
\resizebox{\linewidth}{!}{
\begin{tabular}{cccccccccccc}
\toprule
&&&&& \multicolumn{2}{c}{TLIO Dataset (NN)} & \multicolumn{4}{c}{TLIO Dataset (NN+EKF)}\\
\cmidrule(lr){6-7} \cmidrule(lr){8-11}
ev. & pol. & noise & ref. $\mathbf{x}(t)$ & Manifold & MSE*
& ATE* &  ATE&RTE & Drift & AYE \\
&&&&&($m^2$)&($m$)&($m$)&($m$)&($\%$)&($\circ

$)\\
\midrule
\xmark&-     & -    &-       & -                &\textbf{0.013}&1.660&1.410&\textbf{0.102}&1.486&1.832\\
\cmark&\xmark& -    &position&$\mathbb{R}^3$    &0.014&1.515&1.724&0.110&1.580&1.846\\
\cmark&\xmark& -    &pose    &$SO(3)\times \mathbb{R}^3$&0.015&1.421&1.622&0.108&1.091&1.919\\
\cmark&\xmark& -    &pose    &$SE(3)$           &0.016&1.425&1.420&0.108&0.982&1.862\\
\cmark&\cmark&\xmark&pose    &$SE(3)$           &0.005&\textbf{0.912}&6.080&0.240&3.140&\textbf{1.700}\\
\cmark&\cmark&\cmark&pose    &$SE(3)$           &0.015&1.445&\textbf{1.282}&0.106&\textbf{0.953}&1.863\\

\bottomrule
\end{tabular}}
\caption{ Ablation on event generation, polarity, and (polarity) noise.}
\label{tab:ablation}\vspace{-4ex}
\end{small}
\end{table}

%% file: data_efficiency.tex
\begin{table}[t!]
\centering
\resizebox{0.9\linewidth}{!}{
\begin{tabular}{l|cccc}
\hline
\multicolumn{1}{c|}{\textbf{Testing}} & \multicolumn{4}{c}{\textbf{Training Data}}                                     \\
\multicolumn{1}{c|}{\textbf{Data}}    &  \textbf{50\%} & \textbf{30\%} & \textbf{20\%} & \textbf{10\%} \\ \hline
RoNIN-U &  (5.35/4.63) & (6.11/4.89) & (5.83/4.96) & (5.73/5.00) \\\hline
RIDI-T &  (1.03/1.97) & (1.23/2.50) & (1.22/2.44) & (1.43/2.77) \\\hline
OxIOD & (1.52/1.91) & (1.48/2.82) & (1.52/2.18) & (1.51/2.48)\\ \hline
\end{tabular}}\caption{Data efficiency of our method (ATE$^*$/RTE$^*$) in meters.}\vspace{-6ex}
\label{tab:data_eff}
\end{table}

%% file: 5_conclusion.tex
\section{Limitations and Conclusion}
\noindent\textbf{Limitations:} Pre-integrations as reference signals are efficient to compute, but suffer from slight drift and sampling noise, which may impact event generation. Yet, we show that event generation is robust to sampling noise. Finding better-behaved reference signals is the subject of future work. 

\noindent\textbf{Conclusion:} This work enhances the robustness of neural inertial odometry methods by training the underlying \emph{neural displacement priors} with \textit{Lie events} generated from IMU pre-integrations. We showed both theoretically and empirically that these events have favorable canonicalization properties with respect to motion and rate variations in IMU data which simplifies network training. To generate events on pre-integrations, we generalize notions of level-crossing and event polarities to arbitrary Lie groups, and in particular, the special Euclidean group $SE(3)$. To this end, we introduce new concepts such as \textit{Lie polarities}. Training on these events effectively reduces the tracking error of off-the-shelf neural inertial odometry methods with only minimal input pre-processing. As a result of this study, we believe that many more sensors than IMUs and cameras may benefit from event-based sampling, opening the door to exciting new applications. 

\section*{Acknowledgements}
We gratefully acknowledge support by the following grants: NSF FRR 2220868, NSF IIS-RI 2212433, ONR N00014-22-1-2677 and SNF 225354.

%% file: X_suppl.tex
\setcounter{page}{1}
Here we provide additional material concerning the proof of Theorem I in the main text (Sec.~\ref{sec:app:theorem}), additional details on TLIO (Sec.~\ref{sec:app:tlio}), additional details on RoNIN (Sec.~\ref{sec:app:ronin}), network details (Sec.~\ref{sec:app:network_details}), additional dataset details (Sec.~\ref{sec:app:datasets}), additional metric details (Sec.~\ref{sec:app:metrics}), additional qualitative results (Sec.~\ref{sec:app:qualitative}) and finally result tables for rate and hyperparameter sensitivity studies (Sec.~\ref{sec:app:sensitivity}).
\section{Proof of Theorems}
\label{sec:app:theorem}
\subsection{Proof of Theorem 1}
For completeness, we will restate the theorem here, and then present the proof.

\begin{theorem}
If $\mathbf{x}(t)=\mathbf{x}^*(\phi(t))$ with $\phi'(t)>0$ and \\$\phi(0)=0,\phi(T)=1$, then $\tau_j = \phi^{-1}(\sigma^*_j)$ and $\mathbf{x}_{\text{ref},j}=\mathbf{x}^*_{\text{ref},j}$, where $\tau_j^*$ and $\mathbf{x}^*_{\text{ref},j}$ are generated from $\mathbf{x}^*(t)$.
\end{theorem}

\noindent\textbf{Proof:} We perform induction on $j$, starting with $j=1$. We see that 
\begin{align}
\nonumber\tau_1&=\phi^{-1}(\tau_1)=\phi^{-1}(\sigma^*_1)\\
\nonumber\mathbf{x}_{\text{ref},1}&=\mathbf{x}^*(\phi(\tau_1))=\mathbf{x}^*(\tau_1)=\mathbf{x}^*_{\text{ref},1}    
\end{align}
so, assuming the induction hypothesis for $j-1$, we have 
\begin{align}
   \nonumber\tau_j &= \min_{t>\tau_{j-1}}\{t : \theta \leq \left\Vert \mathbf{x}(t) - \mathbf{x}_{\text{ref},j-1}\right\Vert\}\\
\nonumber&= \min_{t>\tau_{j-1}}\{t : \theta \leq \left\Vert \mathbf{x}^*(\phi(t)) - \mathbf{x}^*_{\text{ref},j-1}\right\Vert\}\\
\nonumber&= \min_{\phi(t)>\phi(\tau_{j-1})}\{t : \theta \leq \left\Vert \mathbf{x}^*(\phi(t)) - \mathbf{x}^*_{\text{ref},j-1}\right\Vert\}\\
\nonumber&= \min_{s>\sigma^*_{j-1}}\{\phi^{-1}(s) : \theta \leq \left\Vert \mathbf{x}^*(s) - \mathbf{x}^*_{\text{ref},j-1}\right\Vert\}\\
\nonumber&= \phi^{-1}\left(\min_{s>\sigma^*_{j-1}}\{s : \theta \leq \left\Vert \mathbf{x}^*(s) - \mathbf{x}^*_{\text{ref},j-1}\right\Vert\}\right)\\
\nonumber&= \phi^{-1}\left(\sigma^*_j\right)
\end{align}
In the first step we use the definition of $\mathbf{x}(t)=\mathbf{x}^*(\phi(t))$. In the second step, we note that $t>\tau_j$ can be replaced with $\phi(t)>\phi(\tau_{j-1})$ since $\phi'(t)>0$. Then, in the third step, we make the substitution $t=\phi^{-1}(s)$ and use the induction hypothesis that $\phi(\tau_j)=\tau_j^*$. Finally, we take out the application of $\phi^{-1}$ and note that the remaining term is exactly the definition of $\sigma^*_j$.
This completes the proof for $\tau_j=\phi^{-1}(\tau_j^*)$. Finally, we see that 
\begin{equation}
    \noindent\mathbf{x}_{\text{ref},j}=\mathbf{x}(\tau_j)=\mathbf{x}^*(\phi(\phi^{-1}(\sigma^*_j)))=\mathbf{x}^*(\sigma^*_j)=\mathbf{x}^*_{\text{ref},j},
\end{equation}
which completes the proof for $\mathbf{x}_{\text{ref},j}=\mathbf{x}^*_{\text{ref},j}$.\qed

\subsection{Extension to Lie Groups}
The proof follows the same steps as before, so we only focus on proving the induction step

\noindent\textbf{Proof:} We have 
\begin{align}
   \nonumber\tau_j &= \min_{t>\tau_{j-1}}\left\{t : \theta \leq \left\Vert \text{Log}(\mathbf{x}^{-1}_{\text{ref},j-1}\mathbf{x}(t))\right\Vert\right\}\\
   \nonumber &= \min_{t>\tau_{j-1}}\left\{t : \theta \leq \left\Vert \text{Log}({\mathbf{x}^*}^{-1}_{\text{ref},j-1}\mathbf{x}^*(\phi(t)))\right\Vert\right\}\\
   \nonumber &= \min_{\phi(t)>\phi(\tau_{j-1})}\left\{t : \theta \leq \left\Vert \text{Log}({\mathbf{x}^*}^{-1}_{\text{ref},j-1}\mathbf{x}^*(\phi(t)))\right\Vert\right\}\\
   \nonumber &= \min_{s>\sigma^*_{j-1}}\left\{\phi^{-1}(s) : \theta \leq \left\Vert \text{Log}({\mathbf{x}^*}^{-1}_{\text{ref},j-1}\mathbf{x}^*(s)\right\Vert\right\}\\
  \nonumber &= \phi^{-1}\left(\min_{s>\sigma^*_{j-1}}\left\{s : \theta \leq \left\Vert \text{Log}({\mathbf{x}^*}^{-1}_{\text{ref},j-1}\mathbf{x}^*(s)\right\Vert\right\}\right)\\
  \nonumber&=\phi^{-1}(\sigma^*_j).
\end{align}
The rest follows similar to Theorem 1.

\section{TLIO Architecture}
\label{sec:app:tlio}
In this section, we discuss in detail the TLIO Architecture~\cite{liu2020tlio} for purely Inertial Odometry with NDP. The goal is to estimate the full pose from a single IMU consisting of accelerometer measuring linear accelerations $\bar{\boldsymbol{a}}(t_i)\in\mathbb{R}^3$, and a gyroscope providing angular velocity measurements $\bar{\boldsymbol{\omega}}(t_i)\in\mathbb{R}^3$ at discrete times $t_i$. The EKF continuously estimates the orientation ($\mathbf{R}(t_i)$), linear velocity (${\mathbf{v}}(t_i)\in\mathbb{R}^3$), position (${\boldsymbol{p}}(t_i)\in\mathbb{R}^3$), accelerometer bias ($\mathbf{b}^a (t_i)\in\mathbb{R}^3$), gyroscope bias ($\mathbf{b}^g (t_i)\in\mathbb{R}^3$). The EKF estimates at the IMU input frequency rate, i.e., at 1 kHz for TLIO and Aria Right Datasets and at 800 Hz for Aria Left Datasets. In the subsections that follow we describe in detail the EKF and NDP treated as a virtual sensor measurement.
%
\subsection{EKF Details}
\textbf{State Definition:}
The dimension of the EKF state is $6n + 15$ and it is composed of the current state\footnote{We omit the subscript $wb$ and $w$.}: 
\begin{equation}
    \mathbf{\hat{s}}_i = (\hat{\mathbf{R}}_i, \hat{\mathbf{v}}_i, \hat{\boldsymbol{t}}_i, \mathbf{\hat{b}}^g _i, \mathbf{\hat{b}}^a _i)\in\mathbb{R}^{15}
\end{equation}
where subscripts now denote the state at time $t_i$, and $n$ past clone states 
\begin{equation}
\boldsymbol{\hat\xi}_j = ({\hat{\mathbf{R}}_j}, \hat{\boldsymbol{t}}_j)\in\mathbb{R}^6    
\end{equation}
where $j\in \{1,2,...,n\}$ and $n\leq i $. Hence, the full state of the EKF is defined as 
\begin{equation}
 \boldsymbol{\hat X}_i = (\boldsymbol{\hat\xi}_{i-n}, ...., \boldsymbol{\hat\xi}_i, \mathbf{\hat s}_i)   
\end{equation}
We introduce the linearized error state $\delta\boldsymbol{\hat X}_i$
\begin{equation}
    \boldsymbol{\hat X}_{i+1}\approx \boldsymbol{\hat X}_i\oplus \delta\boldsymbol{\hat X}_i
\end{equation}
with 
{\small
\begin{align}
\delta\boldsymbol{\hat X}_i&=(\delta\boldsymbol{\hat\xi}_{i-n}, ...., \delta\boldsymbol{\hat\xi}_i, \delta\mathbf{\hat s}_i)  \\
\delta\boldsymbol{\hat\xi}_j &= (\delta\boldsymbol{\hat\theta}_j, \delta\hat{\boldsymbol{t}}_j)\\
\hat{\mathbf{s}}_i &= ({\boldsymbol{\hat \theta}}_i, \delta{\mathbf{\hat v}}_i, \delta{\boldsymbol{\hat t}}_i, \delta{\mathbf{\hat b}^g }_i, \delta{\mathbf{\hat b}^a} _i)
\end{align}
}
Here $\oplus$ denotes addition for all states except for rotation, where $\mathbf{\hat R}_{i+1}=\text{Exp}(\delta\hat{\boldsymbol{\theta}}_i)\mathbf{\hat R}_{i}$. We use covariance matrix $\mathbf{P}_i$ of shape $(6n + 15) \times (6n + 15)$. The clone states corresponding to a 1-second window are included in the filter state as the heading is unobservable, and the NDP processes data from a 1-second window. \\
\noindent\textbf{Preprocessing:}
The raw imu measurements $\bar{\boldsymbol{a}}_i, \bar{\boldsymbol{\omega}}_i$ are related to actual measurements $\boldsymbol{a}_i, \boldsymbol{\omega}_i$ as follows
{\small
\begin{align}
    \boldsymbol{\bar{\omega}}_i &= \boldsymbol{\omega}_i + \mathbf{b}^g_i + \boldsymbol{\eta}^g_i\\
   \quad \bar{\boldsymbol{a}}_i &= \boldsymbol{a}_i - {\mathbf{R}}^{\intercal} _i  \mathbf{g} + \mathbf{b}^a_i + \boldsymbol{\eta}^a_i
\end{align}
}
where $\mathbf{g}$ is the gravity vector in world frame, and $\boldsymbol{\eta}^g_i$ and $\boldsymbol{\eta}^a_i$ are the IMU noises. 

As in previous works~\cite{liu2020tlio} we assume access to a gravity direction and bias estimate, with which IMU measurements are gravity-compensated and bias-corrected: 
\begin{align}
    \boldsymbol{\hat{\omega}}_i &= \hat{\mathbf{R}}_{g,i}(\bar{\boldsymbol{\omega}}_i - \hat{\boldsymbol{b}}^g_i)\\
    \hat{\boldsymbol{a}}_i &= \hat{\mathbf{R}}_{g,i}(\bar{\boldsymbol{a}}_i - \hat{\boldsymbol{b}}^a_i) + \boldsymbol{g}
\end{align}

\noindent\textbf{Process Model:}
The EKF propagates the state with IMU data $\hat{\boldsymbol{a}}_i, \hat{\boldsymbol{\omega}}_i)$ captured over 50 ms via the strap-down inertial kinematics equations as follows
{\small
\begin{gather}
    \hat{\mathbf{R}}_{i+1} = \hat{\mathbf{R}}_i \text{Exp}(\hat{\boldsymbol{\omega}}_i\Delta t)\\
    \hat{\mathbf{v}}_{i+1} = \hat{\mathbf{v}}_i + \hat{\mathbf{R}}_i\hat{\boldsymbol{a}}_i\Delta t\\
    \hat{\mathbf{t}}_{i+1} = \hat{\mathbf{t}}_i + \hat{\mathbf{v}}_i\Delta t + \frac{1}{2}\hat{\mathbf{R}}_i\hat{\boldsymbol{a}}_i\Delta t^2 \\
    \hat{\mathbf{b}}^g_{i+1} = \hat{\mathbf{b}}^g_i\\
    \hat{\mathbf{b}}^a_{i+1} = \hat{\mathbf{b}}^a_i
\end{gather}
}
where $\Delta t = t_{i+1} - t_i$. Note that we assume this for current state $\mathbf{\hat{s}}_{i}$ but assume past states to remain the same, i.e. 
\begin{equation}
\boldsymbol{\hat \xi}_{i+1}= \boldsymbol{\hat \xi}_{i}
\end{equation}
Linearization of the propagation around the current EKF state $\hat{\boldsymbol{s}}_i$ yields
{\small
\begin{align}
\boldsymbol{\hat s}_{i+1} \approx \boldsymbol{A}^s( \mathbf{\hat s}_i) \delta\mathbf{\hat{s}}_i + \boldsymbol{B}^s( \mathbf{\hat s}_i) \boldsymbol{\eta}_i
\end{align}
}
where $\boldsymbol{A}^s( \mathbf{\hat s}_i)$ corresponds to the Jacobian of the propagation equations with respect to the error state, with shape $15\times 15$, and $\boldsymbol{B}^s(\mathbf{\hat s}_i)$ corresponds to the Jacobian of the propagation equations with respect to the noise of shape $15\times 12$
\begin{equation}
\boldsymbol{\eta}_i = [\boldsymbol{\eta}^g_i,\boldsymbol{\eta}^a_i,\boldsymbol{\eta}^{gd}_i, \boldsymbol{\eta}^{ad}_i]^T    
\end{equation}

\textbf{State Augmentation:} The state augmentation is a partial propagation step from the previous IMU timestamp to the new clone state. The error state is first augmented followed by error propagation following the equation below
{\small
\begin{align}
\boldsymbol{P}_{i+1} = \boldsymbol{A}(\mathbf{\hat s}_i) \boldsymbol{P}_i \boldsymbol{A}(\mathbf{\hat s}_i)^T + \boldsymbol{B}(\mathbf{\hat s}_i) \boldsymbol{W} \boldsymbol{B}(\mathbf{\hat s}_i)^T
\end{align}
}
\[
\boldsymbol{A}(\mathbf{\hat s}_i) = \begin{bmatrix}
\boldsymbol{I}_{6n} & 0 \\
0 & \boldsymbol{A}^s(\mathbf{\hat s}_i)
\end{bmatrix},  \quad
\boldsymbol{B}(\mathbf{\hat s}_i) = \begin{bmatrix}
0 \\
\boldsymbol{B}^s(\mathbf{\hat s}_i)
\end{bmatrix},
\]
$\boldsymbol{W}$ corresponds to the noise covariance of the IMU noise and bias drift noise.\\
\noindent\textbf{Measurement Model} 
We assume a measurement model that returns the displacement $\mathbf{d}\in\mathbb{R}^3$ on a 1-second window. Let clone state $\boldsymbol{\hat{\xi}}_j$ be at time $t_j$, the beginning of the window, and let the current state $\mathbf{\hat s}_i$ be at the end of this window. Furthermore, let $\hat{\mathbf{R}}_j = \mathbf{\hat R}_{{\gamma}_j} \mathbf{\hat R}_{{\beta}_j} \mathbf{\hat R}_{{\alpha}_j}$ be the $xyz$ Euler angles of the orientation at $t_j$. As the yaw is unobservable, the NDP and associated covariance are obtained in a local-gravity-aligned frame. Therefore the measurement $h(\mathbf{X})$ is given by the following equation
{\small
\begin{align}
h(\mathbf{X}_i) = \mathbf{R}_{\gamma^T_j} (\boldsymbol{t}_i - \boldsymbol{t}_j)  + \boldsymbol{\eta}^{d}
\end{align}
}
with noise $\boldsymbol{\eta}^d\sim\mathcal{N}(\mathbf{0},\boldsymbol{\Sigma})$. \\
\noindent\textbf{Measurement Update:} We use the displacement and covariance predictions $\mathbf{d},\boldsymbol{\Sigma}$ of the NDP to define the noise covariance of $\boldsymbol{\eta}^d$, and the measurement of displacement. Let $\mathbf{H}_i$ be the Jacobian of the measurement equation with respect to the state with shape $3\times (6n+15)$. The non-zero parts are 
\begin{align}
\mathbf{H}_{\delta{\boldsymbol{\theta}}_i} &= \frac{\partial h(\mathbf{X}_i)}{\partial \delta{\boldsymbol{\theta}}_i} = \hat{\mathbf{R}}_{\gamma,i}^T {\lfloor \hat{\boldsymbol{t}}_j - \hat{\boldsymbol{t}}_i \rfloor}_\times \mathbf{H}_z\\
\boldsymbol{H}_{\delta\tilde{\boldsymbol{p}}_i} &= \frac{\partial h(\mathbf{X})}{\partial \delta\tilde{\boldsymbol{p}}_i} = - \hat{\mathbf{R}}_{\gamma,i}^T\\
\boldsymbol{H}_{\delta\tilde{\boldsymbol{p}}_j} &= \frac{\partial h(\mathbf{X})}{\partial \delta\tilde{\boldsymbol{p}}_j} = \hat{\mathbf{R}}_{\gamma,i}^T
\end{align}
where \[
\boldsymbol{H}_z = \begin{bmatrix}
0 & 0 & 0\\
0 & 0 & 0\\
\cos{\gamma}\tan{\beta} & \sin{\gamma}\tan{\beta}&1
\end{bmatrix}
\] and ${\lfloor \mathbf{x} \rfloor}_\times$ is a skew-symmetric matrix built from a vector $\mathbf{x}$.

Given the measurement Jacobian, let us define the Kalman gain $\mathbf{K}_i$ and innovation $\mathbf{y}_i$ as 
\begin{align}
    \mathbf{K}_i &= \mathbf{P}_i\mathbf{H}^{\intercal}_i\left( \mathbf{H}_i \mathbf{P}_i\mathbf{H}^{\intercal}_i + \boldsymbol{\Sigma}\right)^{-1}\\
    \mathbf{y}_i &= h(\hat{\mathbf{X}}_i) - \mathbf{d}
\end{align}
Then the measurement update equations are 
{\small
\begin{align}
\delta \mathbf{\hat{X}}_i&=\mathbf{K}_i\mathbf{y}_i\\
\mathbf{\hat{X}}_i &= \mathbf{\hat{X}}_i \oplus \delta \mathbf{\hat{X}}_i
\\
\mathbf{P}_i &= (\mathbf{I} - \mathbf{K}_i \mathbf{H}_i) \mathbf{P}_i(\mathbf{I} - \mathbf{K}_i \mathbf{H}_i)^{\intercal} +\mathbf{K}_i\mathbf{\Sigma}\mathbf{K}^{\intercal}_i
\end{align}
}

\section{RoNIN Architecture}
\label{sec:app:ronin}
In this section, we discuss the RoNIN Architecture~\cite{herath2020ronin} which aims for accurate purely inertial odometry with NDP assuming accurate orientation estimates from a smartphone. The goal is to accurately predict 2D velocities $\hat{\boldsymbol{v}}(t_i)\in\mathbb{R}^3$ given fixed 1-second window of gravity-aligned and bias-corrected IMU data $\hat{\boldsymbol{a}}(t_i),\hat{\boldsymbol{\omega}}(t_i)$ where $n$ is number of IMU samples in a 1-second window (i.e., 200 IMU samples for 200Hz IMU data). The IMU data transformed into the gravity-aligned frame using a known orientation estimate and bias-corrected using the factory calibration bias values. RoNIN architecture has three variants as reported in Sec. IV-C Tab.III of the main paper in which the IMU data is temporally processed over sliding windows of IMU samples with a stride of 5 IMU samples using ResNet, Bidirectional LSTM, and TCN. The predicted 2D velocities are integrated to obtain the 2D position estimate $\hat{\boldsymbol{t}}(t_{i+5})$ iteratively as  
\begin{equation}
    \hat{\boldsymbol{t}}(t_{i+5}) = \hat{\boldsymbol{t}}(t_i) + (t_{i+5}- t_i)\hat{\boldsymbol{v}}(t_i)
)\end{equation}

\section{Network Details}
\label{sec:app:network_details}
Here we add network details of RoNIN and TLIO. Both use a 1-D convolutional ResNet-18~\cite{He16cvpr} as encoder. Features after the fourth block are decoded into 2D velocity (RoNIN, one head), or 3D displacement + diagonal covariance (TLIO, two heads). Each head runs a depth-wise 1D Conv+BN block followed by reshaping and a 3-layer MLP with dropout ($p=0.5$) and ReLU non-linearity.  

\section{Dataset Details}
\label{sec:app:datasets}
\textbf{TLIO Dataset:}
We use IMU data from the TLIO test dataset~\cite {liu2020tlio}, which comprises 60 hours worth of 1 kHz IMU data, together with 200 Hz ground truth position, velocity and biases derived from MSCKF~\cite{Mourikis07icra}. It was gathered from five individuals performing a broad range of individual activities, for example, walking, stair traversal, and kitchen organization. It was collected using a custom-built rig with a Bosch BMI055 IMU mounted on a headset attached to cameras. It comprises IMU data sampled at 1 kHz and ground truth comprising 3D positions, orientations, velocities, and biases derived from at 200 Hz. It encompasses 400 sequences, capturing a wide range of pedestrian motion and systematic IMU errors. We follow the data splits in ~\citep{liu2020tlio} allocating 80\% for training, 10\% for validation, and 10\% for testing. 

\textbf{Aria Everyday Dataset:}
The Aria Everyday Dataset~\citep{lv2024aria} is an egocentric dataset collected using the Project Aria glasses~\citep{engel2023project} and comprises a left (800 Hz) and right (1 kHz) IMU together with ground truth position and orientation. It contains 7.3 hours of data, and includes a wide range of wearers engaged in everyday activities like reading, exercising, and relaxing. This dataset encompasses 143 sequences, and it provides two types of ground truth trajectories at 1KHz: a strictly causal open-loop trajectory and a closed-loop trajectory that processes multiple recordings together to represent them into a unified coordinate system. 

\textbf{RoNIN Dataset:}
The RoNIN dataset~\cite{herath2020ronin} comprises 42.7 hours of 200 Hz IMU and ground truth trajectory data collected from pedestrians with varied placements and devices.
Interestingly, IMU placement varied between being in the bag, pocket, or hand of the pedestrian, and also featured IMU's from different devices such as Asus Zenfone AR, Samsung Galaxy S9, and Google Pixel 2XL. However, it must be noted that only 50\% of the collected data is publicly available due to privacy concerns.

\textbf{RIDI Dataset:} The RIDI Dataset~\cite{Yan_2018_ECCV} is a pedestrian dataset collected with 200Hz IMU and ground-truth over 25 hours featuring 10 human subjects. The samples collected capture specific human motion patterns- forward, backward, walking sidewards, acceleration, and deceleration. RIDI dataset does not have accurate orientation estimates and therefore we perform Umeyama alignment~\cite{Umeyama91pami} for a fair comparison with prior work.

\textbf{OxIOD Dataset:} The OxIOD Dataset~\cite{chen2018arxiv} is a smartphone dataset collected in a motion capture system with a 100Hz IMU. The dataset consists of 158 sequences over 42.5 km and 14.72 hours. Interestingly, this dataset features various device placements, and  short distance trajectories.

\section{Metrics}
\label{sec:app:metrics}
Here we write the metric definitions in full detail: \\
\textbf{Mean Squared Error (MSE)}: The mean squared error between predicted ($\hat{\mathbf{d}}(t_i)$) and ground truth displacements ($\mathbf{d}(t_i)$) averaged over the trajectory is defined as
\begin{equation}
\nonumber\text{MSE}=\text{mean}_i \Vert {\mathbf{d}}(t_i) - \hat{\mathbf{d}}(t_i)\Vert^2  
\end{equation}
\noindent\textbf{Absolute Translation Error (ATE)}: It measures the RMSE between estimated ($\bar{\mathbf{t}}(t_i)$) and ground truth positions ($\mathbf{t}(t_i)$), defined as 
\begin{equation}
    \nonumber \text{ATE}=\sqrt{\text{mean}_i \Vert{\mathbf{t}}(t_i) - \bar{\mathbf{t}}(t_i)\Vert^2}
\end{equation}
\textbf{The Relative Translation Error (RTE)}: It compares the local differences between $\mathbf{t}(t_i)$ and $\bar{\mathbf{t}}(t_i)$ over a discrete 1-second window $\delta t$, and is defined as
\begin{equation}
    \text{RTE}=\sqrt{\text{mean}_i \Vert \mathbf{R}_\gamma(t_i)\hat{\mathbf{R}}_\gamma^T(t_i)\Delta\bar{\mathbf{t}}(t_i, \delta t) - \Delta{{\mathbf{t}}}(t_i,\delta t)\Vert^2}
\end{equation}
and $\Delta\bar{\mathbf{t}}(t_i, \delta t)\doteq\bar{\mathbf{t}}(t_i+\delta t) - \bar{\mathbf{t}}(t_i)$
However, in the case of RoNIN where the orientation is not predicted $\mathbf{R}_\gamma\hat{\mathbf{R}}_\gamma^T = I$ and the RTE is calculated over $\delta t$ of 1 minute.\\
\textbf{Absolute Yaw Error (AYE):} It is calculated as 
\begin{equation}
    \text{AYE}=\sqrt{\text{mean}_i \|\gamma(t_i) - \bar{\gamma}(t_i)\|^2}
\end{equation}
\noindent\textbf{Drift:} The drift over the total distance traveled, defined as 
\begin{equation}
\text{Drift}=({\mathbf{t}}(t_N) - \bar{\mathbf{t}}(t_N)) / \sum_i\|{\mathbf{t}}_{i+1} - {\mathbf{t}}(t_i)\|    
\end{equation}

\section{Qualitative Results}
\label{sec:app:qualitative}

Here we show qualitative results of our method on various sequences, visualized in Fig.~\ref{fig:qualitative}. We see that our method reduces drift across all sequences, thus reducing ATE. 
\input{qualitative.tex}

\section{Rate Sensitivity and Hyperparameter Sensitivity}
\label{sec:app:sensitivity}

We present the tabular results of the sensitivity plots presented in Sec. IV-D of the main text in Tab.~\ref{rate_robustness_table}.
\input{robustness_rates}
We also present the tabular results of our hyperparameter sensitivity study (discussed in Sec. IV-E in the main text) for both the neural network alone and in combination with the EKF, trained on the TLIO Dataset. The contrast threshold sensitivity analysis is presented in Tab.~\ref{contrast_threshold}. The sensitivity studies on initial velocity noise and polarity noise are presented in Tab.~\ref{initial_velocity} and Tab.~\ref{polarity_noise}, respectively.
\input{contrast_threshold_sensitivity}
\input{initial_velocity_sensitivity}
\input{polarity_noise_sensitivity}

%% file: qualitative.tex
\begin{figure*}
    \centering
    \includegraphics[width=\linewidth]{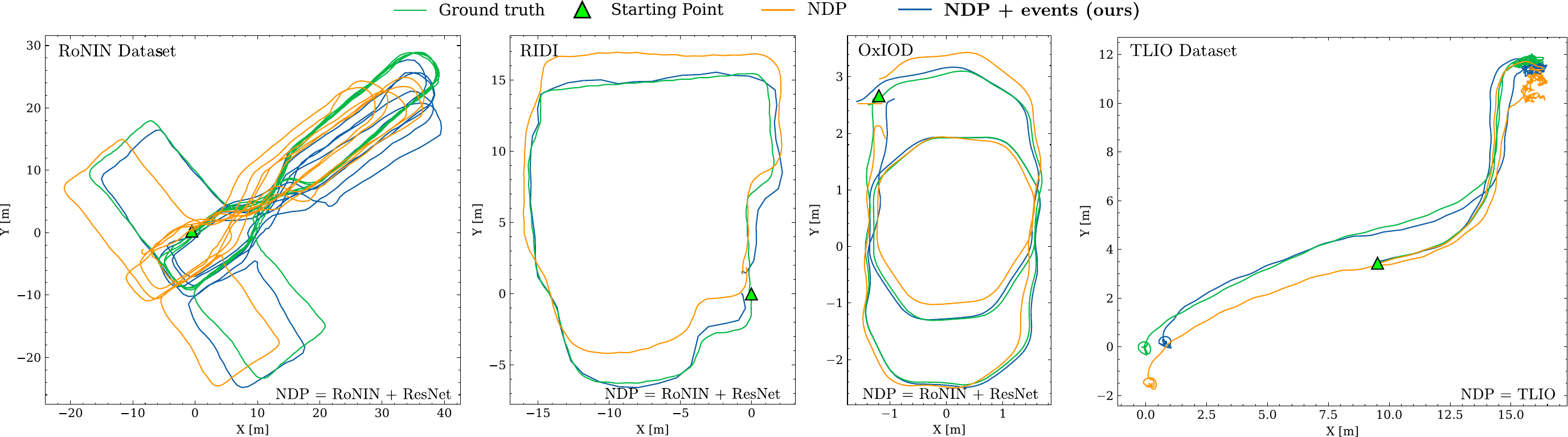}
    \caption{Qualitative results of our method on various sequences. Our method (blue) combines a neural displacement prior (NDP) with Lie events, and reduces overall drift compared to the NDP with regular IMU data as input. The left three sequences use RoNIN + ResNet~\cite{herath2020ronin} as a baseline NDP, and the right sequence uses TLIO~\cite{liu2020tlio}.}
    \label{fig:qualitative}
\end{figure*}

%% file: robustness_rates.tex
\begin{table*}[ht]
\newcommand{\first}{\cellcolor{red!40}}
\newcommand{\second}{\cellcolor{orange!40}}
\newcommand{\third}{\cellcolor{yellow!40}}
\centering
\begin{small}
\resizebox{1.02\linewidth}{!}{
\begin{tabular}{l|c|cccccccccccccccc}
\toprule
&&\multicolumn{4}{c}{TLIO Dataset}&\multicolumn{6}{c}{Aria Right}&\multicolumn{6}{c}{Aria Left}\\
\cmidrule(lr){3-6}\cmidrule(lr){7-12}\cmidrule(lr){13-18}
Model (NN- MSE* $(m^2)$) &rate aug.&
 200 & 100 & 40 & 20 & 1000&500&200& 100 & 40 & 20& 800&400&200& 100 &40 & 20\\
\midrule
TLIO&\xmark&\textbf{0.013}&\textbf{0.013}&0.016&0.032&\underline{0.018}&\underline{0.018}&\underline{0.018}&\underline{0.017}&\underline{0.018}&0.030&0.020&0.020&\underline{0.019}&\underline{0.017}&\underline{0.018}&0.029\\
+ interp.&\cmark&\textbf{0.013}&\textbf{0.013}&\textbf{0.014}&\underline{0.025}&0.019&0.019&0.019&\underline{0.017}&0.019&\underline{0.027}&0.020&0.020&0.020&0.019&\underline{0.018}&\underline{0.028}\\
+ splat.&\cmark&0.014&0.014&\underline{0.015}&15.490&\underline{0.018}&0.019&0.019&0.019&0.019&15.108&\underline{0.018}&\underline{0.018}&0.020&0.019&0.020&14.937\\
+ \textbf{events (ours)}&\xmark&\underline{0.015}&\underline{0.015}&\underline{0.015}&\textbf{0.018}&\textbf{0.015}&\textbf{0.015}&\textbf{0.015}&\textbf{0.015}&\textbf{0.017}&\textbf{0.021}&\textbf{0.015}&\textbf{0.015}&\textbf{0.015}&\textbf{0.015}&\textbf{0.016}&\textbf{0.020}\\
\bottomrule
\toprule
&&\multicolumn{4}{c}{TLIO Dataset}&\multicolumn{6}{c}{Aria Right}&\multicolumn{6}{c}{Aria Left}\\
\cmidrule(lr){3-6}\cmidrule(lr){7-12}\cmidrule(lr){13-18}
Model (NN- ATE* $(m)$) &rate aug.
& 200 & 100 & 40 & 20 & 1000&500&200& 100 & 40 & 20& 800&400&200& 100 &40 & 20\\
\midrule
TLIO&\xmark&1.660&1.612&1.862&3.200&1.283&1.282&1.240&1.164&1.281&1.714&1.432&1.421&1.314&1.228&\underline{1.111}&1.573\\
+ interp.&\cmark&1.519&1.489&\underline{1.565}&\underline{2.440}&1.233&1.217&\underline{1.195}&\underline{1.137}&\underline{1.172}&\underline{1.530}&1.347&1.350&1.326&1.205&1.150&\underline{1.495}\\
+ splat.&\cmark&\underline{1.498}&\underline{1.396}&1.638&192.006&1.216&1.215&1.216&1.281&1.187&156.320&\underline{1.129}&\underline{1.133}&\underline{1.292}&\textbf{1.043}&1.282&170.194\\
+ \textbf{events (ours)}&\xmark&\textbf{1.445}&\textbf{1.386}&\textbf{1.387}&\textbf{1.615}&\textbf{1.077}&\textbf{1.097}&\textbf{1.099}&\textbf{1.103}&\textbf{1.169}&\textbf{1.359}&\textbf{1.054}&\textbf{1.056}&\textbf{1.064}&\underline{1.066}&\textbf{1.087}&\textbf{1.222}\\
\bottomrule
\toprule
&&\multicolumn{4}{c}{TLIO Dataset}&\multicolumn{6}{c}{Aria Right}&\multicolumn{6}{c}{Aria Left}\\
\cmidrule(lr){3-6}\cmidrule(lr){7-12}\cmidrule(lr){13-18}
Model (EKF- ATE $(m)$) &rate aug.
& 200 & 100 & 40 & 20 & 1000&500&200& 100 & 40 & 20& 800&400&200& 100 &40 & 20\\
\midrule
TLIO&\xmark&\underline{1.410}&1.403&1.395&4.282&0.863&0.866&0.865&0.850&0.906&1.253&1.043&1.048&1.069&0.955&0.977&1.255\\
+ interp.&\cmark&\underline{1.410}&\underline{1.336}&\underline{1.353}&\underline{2.906}&\textbf{0.830}&\textbf{0.825}&\underline{0.837}&\textbf{0.823}&\underline{0.889}&\underline{1.150}&\textbf{0.913}&\textbf{0.919}&\underline{0.944}&\textbf{0.922}&\textbf{0.974}&\underline{1.185}\\
+ splat.&\cmark&1.440&1.381&1.471&217.412&0.857&0.857&0.876&0.829&0.893&184.541&0.961&0.967&0.965&0.956&0.996&204.049\\
+ \textbf{events (ours)}&\xmark&\textbf{1.282}&\textbf{1.323}&\textbf{1.335}&\textbf{1.561}&\underline{0.834}&\underline{0.841}&\textbf{0.836}&\underline{0.828}&\textbf{0.852}&\textbf{1.038}&\underline{0.947}&\underline{0.950}&\textbf{0.942}&\underline{0.951}&\underline{0.980}&\textbf{1.035}\\
\bottomrule

\end{tabular}}
\caption{IMU rate sensitivity analysis. Each method is trained on the TLIO training set. Methods + interp. and + splat. were trained with IMU rate augmentation and TLIO and TLIO + events were trained without data rate augmentation. }
\label{rate_robustness_table}
\vspace{-4mm}
\end{small}
\end{table*}

%% file: contrast_threshold_sensitivity.tex
\begin{table}[!t]
\newcommand{\first}{\cellcolor{red!40}}
\newcommand{\second}{\cellcolor{orange!40}}
\newcommand{\third}{\cellcolor{yellow!40}}
\centering
\scriptsize
\begin{small}
\resizebox{\linewidth}{!}{
\begin{tabular}{lccccccc}
\toprule
& \multicolumn{2}{c}{TLIO Dataset (NN)} & \multicolumn{4}{c}{TLIO Dataset (NN+EKF)}\\
\cmidrule(lr){2-3} \cmidrule(lr){4-7}
Contrast& MSE*
& ATE* & ATE & RTE & Drift & AYE \\
Threshold&($m^2$)&($m$)&($m$)&($m$)&($\%$)&($\circ$)\\
\midrule
0.02&0.013&\textbf{1.354}&1.343&0.108&1.108&\textbf{1.807}\\
\textbf{0.01}&0.015&1.445&\textbf{1.282}&\textbf{0.106}&\textbf{0.953}&1.863\\
0.005&\textbf{0.012}&1.592&1.301&0.109&1.209&1.848\\

\bottomrule

\end{tabular}}
\caption{ Contrast Threshold Sensitivity Analysis. The contrast threshold determines the distance to the reference when events are fired, with an optimal value of 0.01.}
\label{contrast_threshold}
\end{small}
\vspace{-4mm}
\end{table}

%% file: initial_velocity_sensitivity.tex
\begin{table}[!t]
\newcommand{\first}{\cellcolor{red!40}}
\newcommand{\second}{\cellcolor{orange!40}}
\newcommand{\third}{\cellcolor{yellow!40}}
\centering
\scriptsize
\begin{small}
\resizebox{\linewidth}{!}{
\begin{tabular}{lccccccc}
\toprule
& \multicolumn{2}{c}{TLIO Dataset (NN)} & \multicolumn{4}{c}{TLIO Dataset (NN+EKF)}\\
\cmidrule(lr){2-3} \cmidrule(lr){4-7}
Initial Velocity& MSE*
& ATE* & ATE & RTE & Drift & AYE \\
Noise Range&($m^2$)&($m$)&($m$)&($m$)&($\%$)&($deg$)\\
\midrule
1.5&0.015&1.561&1.778&0.113&1.074&1.859\\
1.0&0.015&1.485&1.343&0.109&1.114&\textbf{1.820}\\
\textbf{0.5}&0.015&1.445&\textbf{1.282}&\textbf{0.106}&\textbf{0.953}&1.863\\
0.25&\textbf{0.009}&\textbf{1.184}&1.335&0.109&1.076&1.761\\

\bottomrule
\vspace{-3.5mm}
\end{tabular}}
\caption{Initial Velocity Sensitivity Analysis.$\mathbf{v}_0$ noise determines the range of uniform noise perturbing ground truth $\mathbf{v}_0$ used for pre-integration during training. The optimal value of initial velocity noise range is 0.5 to bridge the gap between training with ground-truth and evaluation when combined with an EKF.}
\label{initial_velocity}
\end{small}
\vspace{-4mm}
\end{table}

%% file: polarity_noise_sensitivity.tex
\begin{table}
\newcommand{\first}{\cellcolor{red!40}}
\newcommand{\second}{\cellcolor{orange!40}}
\newcommand{\third}{\cellcolor{yellow!40}}
\centering
\scriptsize
\begin{small}
\resizebox{\linewidth}{!}{
\begin{tabular}{lccccccc}
\toprule
& \multicolumn{2}{c}{TLIO Dataset (NN)} & \multicolumn{4}{c}{TLIO Dataset (NN+EKF)}\\
\cmidrule(lr){2-3} \cmidrule(lr){4-7}
Polarity& MSE*
& ATE* & ATE & RTE & Drift & AYE \\
Noise Range&($m^2$)&($m$)&($m$)&($m$)&($\%$)&($deg$)\\
\midrule
1.0&0.015&1.608&1.292&0.107&\underline{1.048}&\underline{1.836}\\
0.75&0.014&1.477&\textbf{1.219}&\textbf{0.106}&1.107&1.846\\
\textbf{0.5}&0.015&1.445&\underline{1.282}&\textbf{0.106}&\textbf{0.953}&1.863\\
0.25&0.013&1.325&1.484&0.107&1.139&1.879\\
0.1&\underline{0.008}&\underline{1.126}&1.622&0.115&1.060&1.898\\
0&\textbf{0.005}&\textbf{0.912}&6.080&0.240&3.140&\textbf{1.700}\\

\bottomrule
\vspace{-3.5mm}
\end{tabular}}
\caption{Polarity Noise Sensitivity Analysis. The polarity noise determines the noise range added to polarity $\mathbf{p}(\tau_j)$ during training. The optimal value for the polarity noise range is 0.5.}
\label{polarity_noise}
\end{small}
\vspace{-7mm}
\end{table}